%% file: Trajectory_Prediction_Similarity_Based_TXY_RsN1.tex
\documentclass[journal]{IEEEtran}
\usepackage{graphicx}
\usepackage{cite}
\usepackage{picinpar}
\usepackage{amsmath}
\usepackage{url}
\usepackage{flushend}
\usepackage[latin1]{inputenc}
\usepackage{colortbl}
\usepackage{soul}
\usepackage{multirow}
\usepackage{pifont}
\usepackage{xcolor}
\usepackage{alltt}
\usepackage[hidelinks]{hyperref}
\usepackage{enumerate}
\usepackage{siunitx}
\usepackage{breakurl}
\usepackage{epstopdf}
\usepackage{pbox}
\usepackage{amsfonts}
\usepackage{overpic}    
\usepackage{subfigure}  
\usepackage{graphics}
\usepackage{booktabs}
\makeatletter
\newif\if@restonecol
\makeatother

\usepackage[linesnumbered,ruled,vlined]{algorithm2e}
\usepackage{algpseudocode}
\usepackage{amsmath}
\newcommand{\sign}[1]{\mathrm{sgn}(#1)}

\usepackage{courier}
\usepackage{soul}
\soulregister\cite7

\begin{document}
\title{Fast and Accurate Multi-Agent Trajectory Prediction For Crowded Unknown Scenes}

\author{
	\vskip 1em
	
	Xiuye Tao,
	Huiping Li, \emph{Senior Member, IEEE},
	Bin Liang, \emph{Senior Member, IEEE},
	Yang Shi, \emph{Fellow, IEEE}
     and Demin Xu

	\thanks{
	
X. Tao, H. Li and D. Xu are with School of Marine Science and Technology, Northwestern Polytechnical University, Xi'an, 710072, China, (e-mail: taoxiuye@mail.nwpu.edu.cn; lihuiping@nwpu.edu.cn; xudm@nwpu.edu.cn); B. Liang is with Tsinghua University (e-mail: bliang@tsinghua.edu.cn); Y. Shi is with Department of Mechanical Engineering, University of Victoria (e-mail: yshi@uvic.ca).
	}
}

\maketitle
	
\begin{abstract}
This paper studies the problem of multi-agent trajectory prediction in crowded unknown environments. A novel energy function optimization-based framework is proposed to generate prediction trajectories. Firstly, a new energy function is designed for easier optimization. Secondly, an online optimization pipeline for calculating parameters and agents' velocities is developed. In this pipeline, we first design an efficient group division method based on Frechet distance to classify agents online. Then the strategy on decoupling the optimization of velocities and critical parameters in the energy function is developed, where the the slap swarm algorithm and gradient descent algorithms are integrated to solve the optimization problems more efficiently. Thirdly, we propose a similarity-based resample evaluation algorithm to predict agents' optimal goals, defined as the target-moving headings of agents, which effectively extracts hidden information in observed states and avoids learning agents' destinations via the training dataset in advance. Experiments and comparison studies verify the advantages of the proposed method in terms of prediction accuracy and speed.
\end{abstract}

\def\abstractname{Note to Practitioners}
\begin{abstract}
Autonomous robots and vehicles are rapidly integrated into social life and industry, and the scenarios that robots work with multiple people or other moving objects in a crowded environment such as streets and factories will be quite common. The real-time and accurate trajectory prediction of multiple agents (such as people, robots and other vehicles) is a pre-requirement to guarantee efficient motion planning and safe navigation of the robot itself. However, existing methods either require prior information to train models or critical parameters in advance or have insufficient prediction accuracy, which are not suitable for robot online planning. To obtain the accurate predicted trajectories in real-time, we propose a new energy function optimization-based framework to forecast multi-agent trajectories in unknown crowded environments. It utilizes the observed data to infer unknown information without the dataset and optimizes the trajectories very efficiently, which can be adopted for robot motion planning and navigation in unknown crowded environments.
\end{abstract}

\begin{IEEEkeywords}
Trajectory prediction, crowded environments, energy function, Frechet distance, similarity-based resample evaluation algorithm.
\end{IEEEkeywords}


\definecolor{limegreen}{rgb}{0.2, 0.8, 0.2}
\definecolor{forestgreen}{rgb}{0.13, 0.55, 0.13}
\definecolor{greenhtml}{rgb}{0.0, 0.5, 0.0}

\IEEEpeerreviewmaketitle

\section{Introduction}

\IEEEPARstart{P}{redicting} future trajectories of agents is essential in practical applications such as target tracking, robot navigation, and autonomous driving. One of the grand challenges lies in how to estimate motion characteristics of agents in crowded scenes, which depends upon a variety of factors, including self behaviors, goal intents, and interactions with other agents. The accurate and real-time prediction of agents' future trajectories in complex dynamic environments is yet resolved and has received great attention, especially with the development of autonomous driving technology. According to the distinction in modeling approaches, trajectory prediction methods can be divided into physics-based prediction, planning-based prediction, and learning-based prediction methods.

In the physics-based trajectory prediction methods, the constant velocity (CV) model is a common tool to predict agents' future states \cite{Lorente2018,ChenY2022,HU2022}, which is generally implemented in some simple scenarios. Considering that the CV model is often disturbed by agents' random motions, Barrios {\em et al}. \cite{Barrios2015} used the Kalman filter to predict the agent's trajectory with a constant acceleration model. To fully use the information of historical observations, Michael {\em et al}. \cite{Michael2015} utilized the least-squares approximation to extract the pedestrians' principal behavior information, enhancing the accuracy of predicted trajectories. To further improve the prediction accuracy, Pool {\em et al}. \cite{Pool2017} built multiple kinematic models to match the agent's motion and select the optimum model to predict its trajectory. However, this method has poor performance in the long-time prediction. Lee {\em et al}. \cite{Lee2017} utilized the interactive multiple models (IMM), which are built with the coordinated turn motion model and the uniform motion model, to predict human future trajectories. To predict far future trajectories of agents, Xie {\em et al}. \cite{Xie2018} proposed the IMM, including both physics-based and maneuver-based models, in which the probabilities of models in IMM are adjusted to predict the short and long-time trajectories. To avoid collision with moving obstacles when the prediction model is invalid, David {\em et al}. \cite{David2020} studied the confidence-aware motion prediction method, which calculates the confidence of the prediction model by Bayesian belief and then estimates the future trajectory with the belief model.

Besides the physics-based methods, the planning-based approaches for trajectory prediction have also received great attention. The planning-based methods are active prediction, which can address agents' interactions and destinations. Considering the interaction between agents and environments, Aoude {\em et al}. \cite{Aoude2013} utilized the rapidly-exploring random tree to predict feasible trajectories of moving obstacles. Another approach to predict trajectories by considering interactions of agents is the reciprocal velocity obstacles (RVO) method \cite{Berg2011}, which builds the collision avoidance constraints of actions by the geometric interpretation of virtual collision radius and relative velocities, and then calculates the collision-free velocity which closes to the set optimal velocity by the linear programming under the constraints. In particular, Kim {\em et al}. \cite{Kim2015} studied the trajectory prediction problem with uncertain observed states, and they used the RVO to build the motion model, and estimated states by the maximum likelihood estimation and the ensemble Kalman filtering to predict future states with the proposed model. To model agents' interactions based on potential fields, the relations between agents' actions considering collision avoidance are built by the attraction and repulsion of the force according to the social force model in \cite{Helbing1995,Kneidl2013}. Based on this method, Andrey {\em et al}. \cite{Andrey2018} proposed the environment-aware social force model, which takes into account local social interactions to predict multi-agent trajectories. Inspired by the forces between molecules, the Lennard-Jones potential field is used to model the motion trends of a group of agents with interactions in \cite{Boucher2023}. In addition, the flocking algorithms, which simulate swarms' movements of animals by the simple local interaction rules of their conduct cohesion, alignment, and separation based on the states of neighbors, are utilized to predict the complex formation of the group \cite{HuJY2022,QiuH2017}. However, a high-precision dynamic model is particularly important for accurate trajectory prediction in complex and crowded environments. To improve the performance of trajectory prediction in crowded scenes by potential fields, the energy-based model, which is built with agents' motion characteristics, destinations, group attractions, and interactions, is proposed to estimate future optimal velocities \cite{Yamaguchi2011,Pellegrini2009}.

The learning-based methods of trajectory prediction have made great achievements. To overcome the weakness of simple interaction models built by handcrafted functions in crowded settings, Vemula {\em et al}. \cite{Vemula2017} utilized the Gaussian processes (GP) to model agents' interactions and trained the GP model with real human trajectory data to predict their future trajectories. Habibi and How proposed a similarity-based multi-model fusion method to predict trajectories, which selects similar trajectories by comparing their motion primitives and then fuses selected trajectories by the GP method to train the prediction model \cite{Golnaz2021}. To fully take advantage of social etiquette, Robicquet {\em et al}. \cite{Alexandre2016} used datasets to train the energy-based model and sorted the optimal parameters into several classes by support vector machine (SVM) offline, then predicted human trajectories online by matching the optimal parameters. To associate relationships of observed states, the Long-Short-Term-Memory (LSTM) network, which models the dependencies of long sequences, has been employed to predict agents' trajectories \cite{Gers2000,Benjamin2016,WangH2021}. To further improve the prediction accuracy, Alexandre {\em et al}. \cite{Alexandre2016cvpr} studied the social LSTM network, which uses the social pooling layer to share hidden states with each other in LSTM. To handle the low computational efficiency of LSTM, Agrim {\em et al}. \cite{Agrim2018} proposed the social Generative Adversarial Networks (GANs) to learn multiple feasible trajectories for human beings.

Though great progress has been made in trajectory prediction, the real-time and accurate prediction for multiple agents in unknown crowded environments is still not available, which is pre-requirement for autonomous robot navigation. In particular, the physics-based method is of low accuracy in comparison with the learning-based method. However, the learning-based method (such as \cite{Golnaz2021, Alexandre2016cvpr, Agrim2018}) needs datasets to train the network, which might not be suitable for robot navigation in unknown environments. Although the planning-based method in \cite{Yamaguchi2011} can provide accurate prediction, it still needs to learn critical parameters in advance. In addition, the computation in \cite{Yamaguchi2011} is too heavy to be implemented online. In this paper, we study the multi-agent trajectory prediction problem in unknown crowded environments without prior information, aiming to develop a real-time and accurate trajectory prediction method for robot navigation in complex unknown environments.

The main contribution of this paper is to develop a novel long-sequence multi-agent trajectory prediction method in unknown crowded environments. The new prediction method is proposed in the framework of a planning-based approach without needing datasets, where a more efficient energy function and optimization scheme have been designed.

In summary, the main contributions of this article are as follows.
\begin{itemize}
\item An efficient group division method based on the Frechet distance is proposed to classify the agent group online, which avoids learning the group pattern offline from using lots of data \cite{Yamaguchi2011}.
\item Strategies are taken to enhance the optimization performance. Firstly, a new energy function modeling the multi-agent trajectory prediction proposed and it is more efficient for optimization. Secondly, the decoupling strategy is designed for the coupled multi-agent trajectories prediction model, which further speeds up optimization. Besides, the salp swarm algorithm (SSA) combined with the gradient descent (GD) method is utilized to facilitate the global optima seeking of model parameters.
\item  The similarity-based resample evaluation (SRE) method is developed to estimate the target heading of each agent, which resolves the issue that the destination of each agent needs to be learned or obtained offline.
\end{itemize}
In addition, comprehensive and comparison studies are carried out to verify the effectiveness and feasibility of the proposed method.

The rest of this article is organized as follows. In Section \ref{Sect2}, the problem formulation is presented. Section \ref{Sec_FrameDesigned_new} designs the framework to optimize parameters of the energy function. In Section \ref{Sec3_group}, the group division by the Frechet distance is proposed to classify observed agents into corresponding groups. Section \ref{Variabletrain} proposes the decoupled cost function to estimate the optimal parameters of the energy function. In Section \ref{S_OptSpeedEst}, we combine the SSA and the GD to calculate agents' optimal velocities under the given parameter sets $\Theta$. The target heading estimation method is designed in Section \ref{Sec4_goal_traj}, and the algorithm of the proposed trajectory prediction is presented. In Section \ref{Sec5_results}, the experiment and comparison are conducted. Finally, Section \ref{Sec6_conclusion} concludes this work.

\section{Background}\label{Sect2}

\subsection{Problem formulation}
Consider the trajectory prediction problems for multiple agents moving in crowded environments. Firstly, for each agent $i$, define state in the navigation coordinate as ${\bf{p}}_t^i = [x_t^i,y_t^i]^{\rm T}$ at time $t$, and define all the observed states of agent $i$ from $t-{\rm N}+1$ to $t$ as ${{\mathbf P}_{t}^i}$, where $\rm N \ge 2$ is the observed length of agent $i$, ${\mathbf P}_{t}^i = [{\bf{p}}_{t - {\rm N} +1}^i,...,{\bf{p}}_{t-2}^i,{\bf{p}}_{t-1}^i,{\bf{p}}_{t}^i]$. In this paper, we aim at predicting all agents' future states by using the observed data $\{{\mathbf P}_{t}^i\}$, $i  \in [1,2,...,{\rm M}_t]$ and ${\rm M}_t$ is the number of observed agents. However, the unknown destinations and the complex interactive behaviors make the complete trajectory prediction intractable, especially in crowded scenes.


In this paper, the energy function-based framework is utilized to predict agents' future states using observed states. An energy function $\mathrm E_{\Theta}(\cdot)$ in \cite{Yamaguchi2011} is formulated as follows:
\begin{equation}
\begin{array}{l}
\vspace{1ex}
{{\rm{E}}_{{\Theta}_i} }({\bf{v}}_{t + k+1}^i|{\bf{p}}_{t + k}^i,\! \{ {\bf{p}}_{t + k}^j\} ) \!=\! {\lambda _0^i}{{\rm{E}}_{damp}}({\bf{v}}_{t + k+1}^i|{\bf{p}}_{t + k}^i) \\
\vspace{1ex}
+{\lambda _1^i}{{\rm{E}}_{speed}}({\bf{v}}_{t + k+1}^i|{\bf{p}}_{t + k}^i) \\
\vspace{1ex}
+ {\lambda _2^i}{{\rm{E}}_{direction}}({\bf{v}}_{t + k+1}^i|{\bf{p}}_{t + k}^i)  \\
\vspace{1ex}
+{\lambda _3^i}{{\rm{E}}_{attraction}}({\bf{v}}_{t + k+1}^i|{{\boldsymbol s}_l})\\
 + {\lambda _4^i}{{\rm{E}}_{group}}({\bf{v}}_{t + k+1}^i|{{\boldsymbol s}_l})  \\
 \vspace{1ex}
+{{\rm{E}}_{collision}}({\bf{v}}_{t + k+1}^i|{\bf{p}}_{t + k}^i,\{ {\bf{p}}_{t + k}^j\} ,{\sigma _d^i},{\sigma _w^i},\beta^i ).
\end{array}
\label{E1_energyfunction}
\end{equation}
The detailed definitions of the energy function are as follows:
\begin{itemize}
\item ${{\rm{E}}_{damp}(\cdot)}$ represents the dampling of sudden changes in the velocity.
\item  ${\rm{E}}_{speed}$ is the penalty term of deviating from the desired speed.
\item ${\rm{E}}_{direction}$ represents the magnitude of the deviation of the velocity dircetion from the destination direction. 
\item ${\rm{E}}_{attraction}$ is the attractiveness of the group to the agent.
\item ${\rm{E}}_{group}$ represents the penalty term of deviating from the group speed.
\item ${\rm{E}}_{collision}$ is the collision avoidance penalty term.
\end{itemize}

\begin{figure}[htb]
	\centering
	\includegraphics[width=8.9cm]{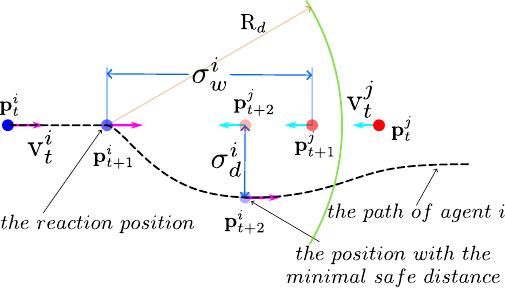}	 
	\caption{Interaction parameters of agent $i$. ${\rm v}_{t}^i$ and ${\rm v}_{t}^j$ are velocities of agents $i$ and $j$, respectively. ${\rm R}_d$ is the detection distance of the sensor on agent $i$, and ${\rm R}_d \ge {\sigma _w^i}$. When agent $i$ moves from ${\bf{p}}_{t}^i$ to ${\bf{p}}_{t+1}^i$, the distance between agents $i$ and $j$ is ${\sigma _w^i}$. To avoid the collision, agent $i$ needs to change its motions. At time $t+2$, agent $i$ maintains the minimal distance ${\sigma _d^i}$ from $j$ to keep its safety.} 
	\label{F1_InteractSigma}
\end{figure}

In \eqref{E1_energyfunction}, $k \in [-{\rm N}+1,-{\rm N}+2,...,0,...,{\rm N}_p ]$, ${\rm N}_p $ represents the predicted length of agents' future trajectories, ${\bf{v}}_{t + k+1}^i $ is the selected velocity vector of agent $i$ in the navigation coordinate, $\{{\bf{p}}_{t+k}^j\} $ are positions of other agents at time $t+k$, and ${\boldsymbol s}_l$ indicates positions of the $l$th group agent $i$ belonging to, ${\Theta}_i = \{ \lambda _0^i,\lambda _1^i,\lambda _2^i,\lambda _3^i,\lambda _4^i,{\sigma _d^i},{\sigma _w^i},\beta ^i\}$ represents the set of parameters for the energy function, in which $\{ \lambda _0^i,\lambda _1^i,\lambda _2^i,\lambda _3^i,\lambda _4^i \}$ are importance weights of sub-cost functions, and ${\sigma _d^i}$ is the minimal distance of agent $i$ with respect to other agents to keep its safety, ${\sigma _w^i}$ represents the maximum distance that agent $i$ reacts to avoid the collision when interacts with other agents, and $\beta^i $ is a weight parameter. The detailed illustrations of interaction parameters ${\sigma _w^i}$ and ${\sigma _d^i}$ are shown in \mbox{Fig. \ref{F1_InteractSigma}}.

The energy function \eqref{E1_energyfunction} is more robust and useful than single prediction styles, such as the physics-based and the RVO models, because it combines multiple factors that primarily affect the agents' motions. According to the characteristics of the energy function, the sub-functions of ${{\rm{E}}_{damp}}(\cdot)$, ${{\rm{E}}_{speed}}(\cdot)$, ${{\rm{E}}_{direction}}(\cdot)$, ${{\rm{E}}_{attraction}}(\cdot)$, ${{\rm{E}}_{group}}(\cdot)$ and ${{\rm{E}}_{collision}}(\cdot)$ in \eqref{E1_energyfunction} can be divided into three categories:

{\bf{a) Self influence to}} ${\bf{v}}_{t + k+1}^i$. This category includes the cost functions of ${{\rm{E}}_{damp}}(\cdot)$, ${{\rm{E}}_{speed}}(\cdot)$ and ${{\rm{E}}_{direction}}(\cdot)$, which are decided by the agent's self-motion behavior. Especially, in the uncrowded environment, they mainly control agent's future paths. The three cost functions are as follows:
\begin{equation}
{{\rm{E}}_{damp}}({\bf{v}}_{t + k+1}^i|{\bf{p}}_{t + k}^i) = {\left\| {{{\bf{v}}_{t + k+1}^i - {\bf{v}}_{t + k}^i}} \right\|}^2,
\label{E2_dampfunction}
\end{equation}

\begin{equation}
{{\rm{E}}_{speed}}({\bf{v}}_{t + k+1}^i|{\bf{p}}_{t + k}^i) = {\left( {\left\|  {{\bf{v}}_{t + k+1}^i} \right\| - {u_i}} \right)}^2,
\label{E3_speedfunction}
\end{equation}

\begin{equation}
{{\rm{E}}_{direction}}({\bf{v}}_{t + k+1}^i|{\bf{p}}_{t + k}^i) = - \frac{({{\bf{z}}_i} - {\bf{p}}_{t + k}^i)^{\rm T}}{{\|{{\bf{z}}_i} - {\bf{p}}_{t + k}^i\|}} \cdot \frac{{{\bf{v}}_{t + k+1}^i}}{{\|{\bf{v}}_{t + k+1}^i\|}},
\label{E4_destinationfunction}
\end{equation}
where $u_i \ge 0$ is the desired speed value of agent $i$, ${{\bf{z}}_i}$ is agent's destination.

{\bf{b) Group influence to}} ${\bf{v}}_{t + k+1}^i$. This includes two terms:
\begin{equation}
\begin{array}{l}
{{\rm{E}}_{attraction}}({\bf{v}}_{t +k+ 1}^i|{{\boldsymbol s}_l}) =  \sum\limits_{j} {\left( {\frac{{({\bf{v}}_{t + k}^i)}^{\rm T}}{{\|{\bf{v}}_{t + k}^i\|}} \cdot \frac{{{\bf{v}}_{t + k}^j}}{{\|{\bf{v}}_{t + k}^j\|}}} \right)}\\
 \cdot \left( {\frac{({\Delta {\bf{p}}_{t + k}^{i,j}})^{\rm T}}{{\|\Delta {\bf{p}}_{t + k}^{i,j}\|}} \cdot \frac{{{\bf{v}}_{t +k+ 1}^i}}{{\|{\bf{v}}_{t + k+1}^i\|}}} \right), \quad s.t.  \quad j \in G_l,\quad j \ne i,
\end{array}
\label{E5_attractfunction}
\end{equation}

\begin{equation}
{{\rm{E}}_{group}}({\bf{v}}_{t +k+ 1}^i|{{\boldsymbol s}_l}) = \left( {\|{\bf{v}}_{t + k+1}^i\| - {u_l}} \right)^2,
\label{E6_groupfunction}
\end{equation}
where ${{\bf{p}}_{t+k}^i},{{\bf{p}}_{t+k}^j} \in{{\boldsymbol s}_l}$, $\Delta {\bf{p}}_{t+k}^{i,j} = {\bf{p}}_{t+k}^{i}-{\bf{p}}_{t+k}^{j}$, $G_l$ represents the index set of agents in the group ${\boldsymbol s}_l$, and ${u_l}$ is the average speed value of the group $l$. In this paper, the group speed $u_l$ is calculated by
\begin{equation}
u_l = \frac{{\sum\limits_{j \in {G_l}} {{u_j}} }}{{\sum\limits_{j \in G_l} 1 }}.
\label{E7new_calculatedGp}
\end{equation}

{\bf{c) Interaction influence to}} ${\bf{v}}_{t +k+ 1}^i$. This term shows how surrounding agents and the environment affect the target agent's action to avoid the potential collision. The interaction cost function is 
\begin{equation}
\begin{array}{ll}
\vspace{1ex}
&{{\rm{E}}_{collision}}({\bf{v}}_{t + k+1}^i|{\bf{p}}_{t + k}^i,\{ {\bf{p}}_{t + k}^j\} ,{\sigma _d^i},{\sigma _w^i},\beta^i ) = \\
\vspace{1ex}
&\sum\limits_{j \ne i} {w({{\bf{p}}_{t + k}^i},{{\bf{p}}_{t + k}^j})\exp \left( { - \frac{{{d^2}({\bf{v}}_{t + k+1}^i,{{\bf{p}}_{t + k}^i},{{\bf{p}}_{t + k}^j})}}{{2{\sigma _d^i}^2}}} \right)} ,
\end{array}
\label{E7_interactionfunction}
\end{equation}
where the cost value $w({{\bf{p}}_{t + k}^i},{{\bf{p}}_{t + k}^j})$ is composed of interactive distances, $w({{\bf{p}}_{t + k}^i},{{\bf{p}}_{t + k}^j})$ and ${d}({\bf{v}}_{t + k+1}^i,{{\bf{p}}_{t + k}^i},{{\bf{p}}_{t + k}^j})$ are
\begin{equation}
\begin{array}{ll}
w({{\bf{p}}_{t + k}^i},{{\bf{p}}_{t + k}^j}) &=\exp \left( { - \frac{{\|\Delta {\bf{p}}_{t + k}^{i,j}{\|^2}}}{{2{\sigma _w^i}^2}}} \right) \\
&\cdot {\left({\frac{1}{2}\left( {1 - \frac{({\Delta {\bf{p}}_{t+k}^{i,j}})^{\rm T}}{{\|\Delta {\bf{p}}_{t+k}^{i,j}\|}} \cdot \frac{{{\bf{v}}_{t+k+1}^i}}{{\|{\bf{v}}_{t+k+1}^i\|}}} \right)}\right)^\beta },
\end{array}
\label{E8_wfunction}
\end{equation}

\begin{equation}
\begin{array}{l}
\vspace{1ex}
{d}({\bf{v}}_{t + k+1}^i,{{\bf{p}}_{t+k}^i},{{\bf{p}}_{t+k}^j}) = \\
{\left\| {\Delta {\bf{p}}_{t+k}^{i,j} \!-\! \frac{{(\Delta {\bf{p}}_{t+k}^{i,j}) ^{\rm T}\cdot ({\bf{v}}_{t + k+1}^i \!-\! {\bf{v}}_{t+k}^j)}}{{\|{\bf{v}}_{t + k+1}^i \!-\! {\bf{v}}_{t+k}^j\|}}({\bf{v}}_{t +k+ 1}^i - {\bf{v}}_{t+k}^j)} \right\|}.
\end{array}
\label{E9_dfunction}
\end{equation}

\section{Optimization Framework Design}\label{Sec_FrameDesigned_new}
\begin{figure}[htb]
	\centering
   \subfigure[]
   {
	\includegraphics[width=7cm]{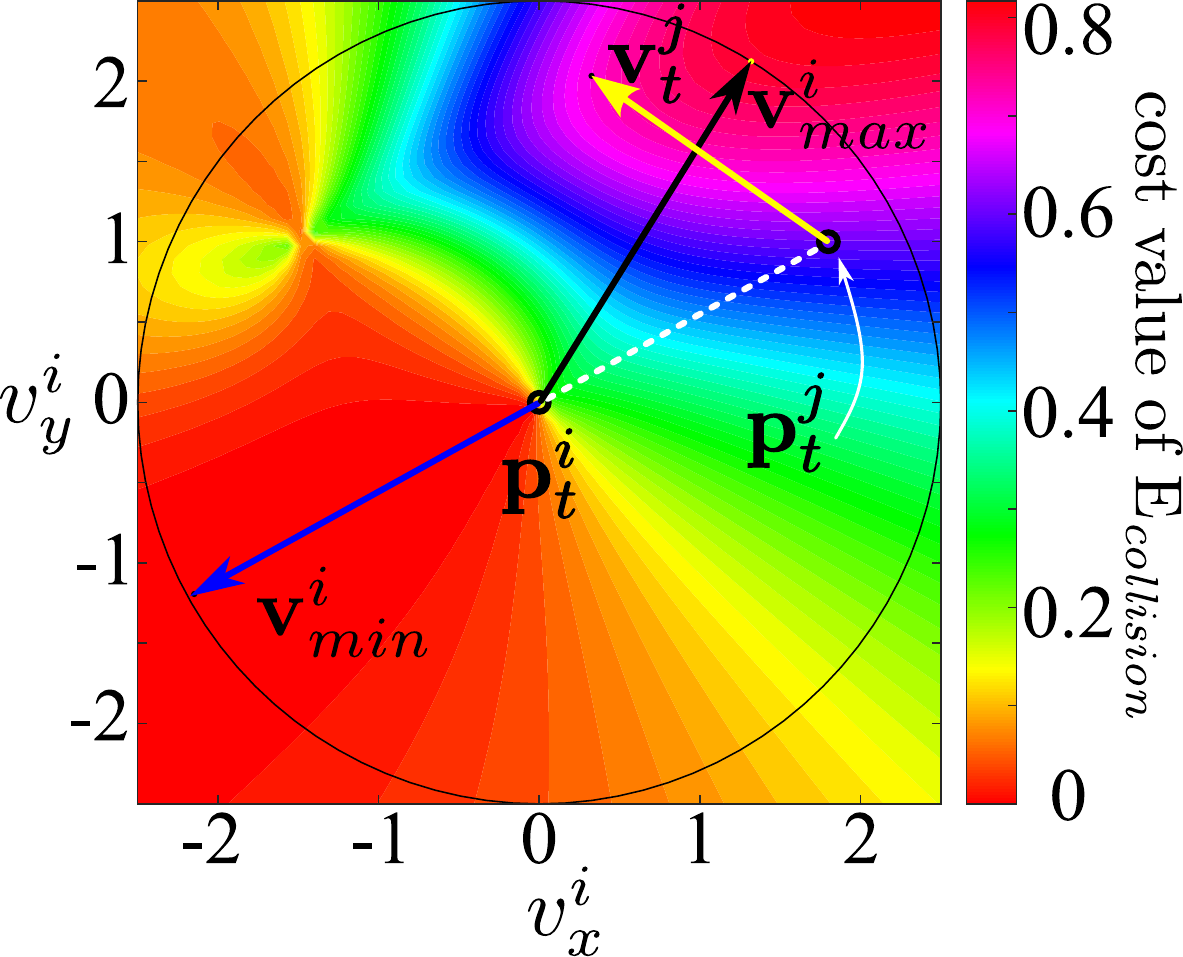}
	\label{F1_Einter_sigma}
   }
   \quad
	\subfigure[]
	{
	\includegraphics[width=7cm]{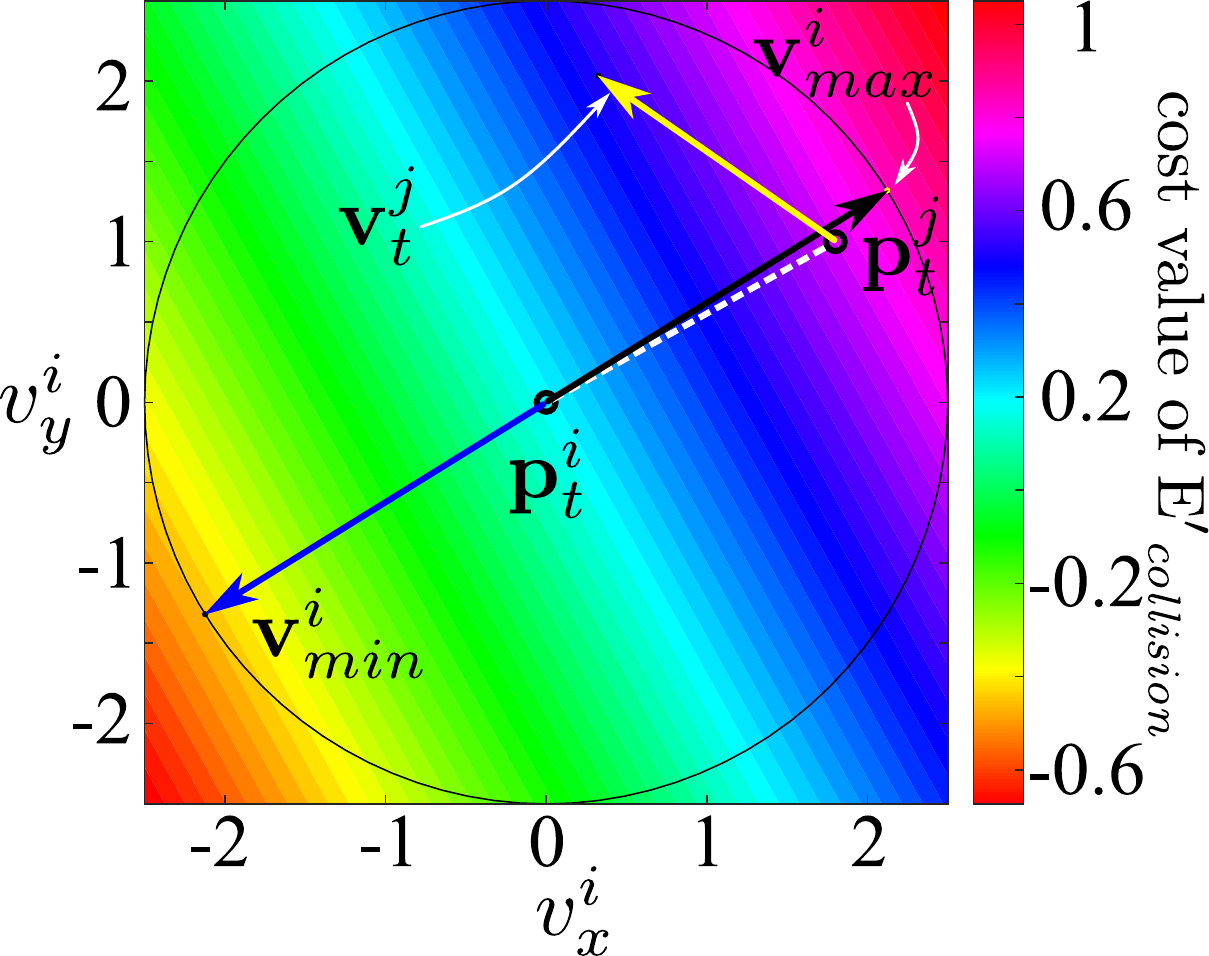}
	\label{F2_Einter_md}
	} 
	\caption{Cost distributions of interaction functions with selecting velocity ${\bf{v}}_{t + 1}^i$. ${\bf{p}}_t^i$ and ${\bf{p}}_t^j$ are positions of agents $i$ and $j$ at time $t$, the yellow arrow lines represent the velocity of agent $j$, the black circles are the velocity constraint of ${\bf{v}}^i$, which is $\left\| {{{\bf{v}}^i}} \right\| \leq 2.5$, the blue arrow lines ${\bf{v}}_{min}^i$ represent selected velocities ${\bf{v}}_{t + 1}^i$, which correspond to minimum values of ${\rm{E}}_{collision}(\cdot)$ and ${\rm{E'}}_{collision}(\cdot)$, the black arrow lines ${\bf{v}}_{max}^i$ are selected velocities ${\bf{v}}_{t + 1}^i$, which correspond to maximum values of ${\rm{E}}_{collision}(\cdot)$ and ${\rm{E'}}_{collision}(\cdot)$. (a) Interaction function ${\rm{E}}_{collision}(\cdot)$ with parameters ${\sigma _d^i}$, ${\sigma _w^i}$, and $\beta^i$. (b) Interaction function ${\rm{E'}}_{collision}(\cdot)$ with parameters ${{\rm{w}}^i}$, ${{\rm d}^i}$, and ${\alpha ^i}$.}
\end{figure}

The interaction function ${\rm{E}}_{collision}(\cdot)$ illustrates that when agent $i$ selects the action ${\bf{v}}_{t +k+ 1}^i$ to make agent $i$ approach to agent $j$, the value of ${\rm{E}}_{collision}(\cdot)$ will grow up, as shown in \mbox{Fig. \ref{F1_Einter_sigma}}. However, the function is too complex, which will affect the optimization speed and reduce the prediction accuracy of actions due to insufficient optimization. Without losing the interaction characteristics of agents, motivated by \cite{Helou2016} we design a new interaction function ${\rm{E'}}_{collision}(\cdot)$ as follows:
\begin{equation}
\begin{array}{ll}
\vspace{1ex}
&{{\rm{E'}}_{collision}}({\bf{v}}_{t + k+1}^i|{\bf{p}}_{t + k}^i,\{ {\bf{p}}_{t + k}^j\} ,{{\rm{w}}^i},{{\rm d}^i},{\alpha^i}) = \\
\vspace{1ex}
&\sum\limits_{j \ne i} {{\rm{D(}}\Delta {\bf{p}}_{t + k}^{i,j}{\rm{)}} \cdot \frac{({\Delta {\bf{p}}_{t + k}^{i,j}})^{\rm T}}{{\|\Delta {\bf{p}}_{t + k}^{i,j}\|}} \cdot \left( {{\bf{v}}_{t + k}^j - {\bf{v}}_{t + k+1}^i} \right)},
\label{E10_Einterdw}
\end{array}
\end{equation}

\begin{equation}
\begin{array}{ll}
&{\rm{D(}}\Delta {\bf{p}}_{t + k}^{i,j}{\rm{)}} = \frac{{{{\rm{w}}^i}}}{{2{{\rm d}^i}}} \cdot \\
&\left( {{{\rm d}^i} - \|\Delta {\bf{p}}_{t + k}^{i,j}\| + \sqrt {{{({{\rm d}^i} - \|\Delta {\bf{p}}_{t + k}^{i,j}\|)}^2} + {\alpha ^i}} } \right),
\end{array}
\label{E11_interbehav}
\end{equation}
where ${\rm{D(}}\Delta {\bf{p}}_{t + k}^{i,j}{\rm{)}}$ represents the influence of interaction distance $ \|\Delta {\bf{p}}_{t + k}^{i,j}\| $ to the motion of agent $i$, ${\rm{w}}^i >0$ is the strength of other agents affecting agent $i$, ${\rm d}^i$ is the interaction distance, and $\alpha ^i$ is the parameter to smoothen the change of ${\rm{D(}}\Delta {\bf{p}}_{t+k}^{i,j}{\rm{)}}$, $\alpha ^i \in [0,{\rm d}^i)$.

The function \eqref{E11_interbehav} shows that when the value of $\|\Delta {\bf{p}}_{t+k}^{i,j}\|$ is in the range of $[0,{\rm d}^i]$, the value of ${\rm{D(}}\Delta {\bf{p}}_{t+k}^{i,j}{\rm{)}}$ will linearly change with $\|\Delta {\bf{p}}_{t+k}^{i,j}\|$, in the case of $\alpha ^i = 0$. Otherwise, other agents have no influence on the motion of agent $i$. 

The cost value in \eqref{E10_Einterdw} with selected actions ${\bf{v}}_{t +k+ 1}^i$ is shown in \mbox{Fig. \ref{F2_Einter_md}}, and it can be seen that the cost values increase along the direction from ${\bf{p}}_{t+k}^i$ to ${\bf{p}}_{t+k}^j$. Compared with \mbox{Fig. \ref{F1_Einter_sigma}}, the value change of \eqref{E10_Einterdw} is smoother, and the maximum values of \eqref{E10_Einterdw} and \eqref{E7_interactionfunction} are mainly concentrated in the upper right corner. There are similar effects of agents' interactions in \eqref{E10_Einterdw} and \eqref{E7_interactionfunction}, however, the function \eqref{E10_Einterdw} is more concise than \eqref{E7_interactionfunction}. To enhance the optimization speed and the accuracy of the selected velocity, \eqref{E10_Einterdw} is utilized to substitute the interaction function in \eqref{E1_energyfunction}, and the new energy function is
\begin{subequations}
\begin{gather}
\begin{array}{ll}
\vspace{1ex}
{\rm{E'}}_{{\Theta _i}}(\cdot) \!= \!&{\lambda _0^i}{{\rm{E}}_{damp}}(\cdot) \!+ {\lambda _1^i}{{\rm{E}}_{speed}}(\cdot) +\\
\vspace{1ex}
&{\lambda _2^i}{{\rm{E}}_{direction}}(\cdot)  +{\lambda _3^i}{{\rm{E}}_{attraction}}(\cdot) +\\
\vspace{1ex}
&{\lambda _4^i}{{\rm{E}}_{group}}(\cdot) +{\rm{E'}}_{collision}(\cdot),
\end{array}
\label{E12_newEnergy}\\
{\Theta}_i = \{ \lambda _0^i,\lambda _1^i,\lambda _2^i,\lambda _3^i,\lambda _4^i,{\rm{w}}^i,{\rm d}^i,\alpha^i \}.
\label{E12_newvariable}
\end{gather}
\end{subequations}

With the energy function \eqref{E12_newEnergy}, the position update of agent $i$ is defined as
\begin{subequations}
\begin{align}
&{\bf{p}}_{t +k+ 1}^i = {\bf{p}}_{t+k}^i + {\bf{v}}_{t + k+1}^{{i_*}}\Delta t,\label{F13_posupdate}\\
s.t. \, &{\bf{v}}_{t +k+ 1}^{{i_*}} = \mathop {\arg \min }\limits_{{\bf{v}}_{t + k+1}^i} {{\rm{E'}}_{\Theta _i^*}}\left( {{\bf{v}}_{t +k+ 1}^i|{\bf{p}}_{t+k}^i,\{ {\bf{p}}_{t+k}^j\} } \right),\label{F14_optV}
\end{align}
\end{subequations}
where $\Delta t$ is the sample interval, $\Theta _i^*$ is the set of the optimal parameters in \eqref{E12_newvariable}, and ${\bf{v}}_{t +k+ 1}^{{i_*}}$ represents the optimal predicted velocity of agent $i$. Referring to \cite{Yamaguchi2011}, the parameters in \eqref{E12_newvariable} are learned with observed data ${{\mathbf P}_t^i}$ by
\begin{equation}
{\Theta ^*} = \mathop {\arg \min }\limits_\Theta  \sum\limits_i {\sum\limits_{k = -{\rm N}+2} {\left\| {{\bf{p}}_{t +k}^i - {\bf{\bar p}}_{t+k}^i} \right\|} } ,
\label{E15_trainpara}
\end{equation}
where $\Theta = [\Theta_1,\Theta_2,...,\Theta_{{\rm M}_t}]$, $k \in [-{\rm N}+2,0]$, ${{\bf{p}}_{t+k}^i}$ denotes the ground truth, and ${\bf{\bar p}}_{t +k}^i = {{\bf{\bar p}}_{t  +k}^i(\bar {\bf p}_{t  +k - 1}^i,\{{\bf p}_{t  +k - 1}^j\} ,{\Theta _i})}$ is the predicted position by \eqref{F13_posupdate} and \eqref{F14_optV} based on the state $\bar {\bf p}_{t  +k - 1}^i$, observed state $\{ {\bf p}_{t  +k - 1}^j\}$ and the given parameters in ${\Theta _i}$.

\begin{figure}[htb]
	\centering
	\includegraphics[width=8.5cm]{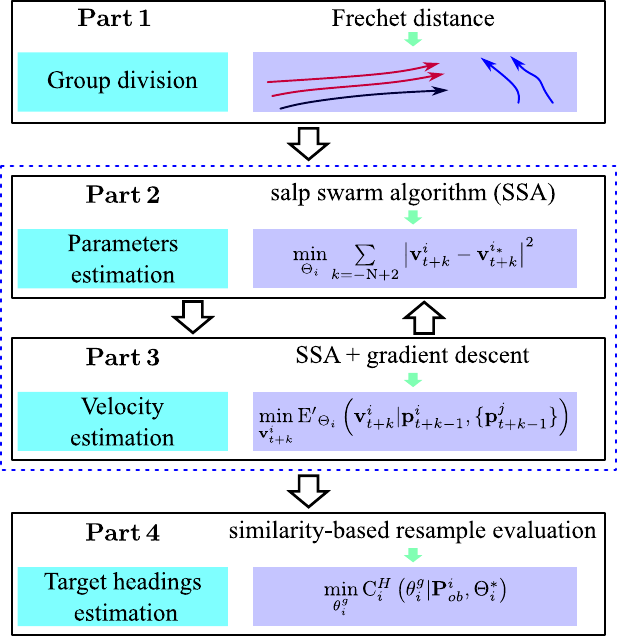}	 
	\caption{Scheme of the proposed trajectory prediction method.} 
	\label{F4_trajPredprocess}
\end{figure}

Note that it is generally impossible to solve \eqref{E15_trainpara} online in crowded scenes directly. Therefore, we need to simplify \eqref{E15_trainpara} and find a way to calculate its approximate optimal solution. Based on the aforementioned problem formulation, the proposed online trajectory prediction is divided into four key parts presented in \mbox{Fig. \ref{F4_trajPredprocess}}.

\section{Group Division}\label{Sec3_group} 
In this section, we first introduce the advantages of the Frechet distance. Then we present the way to divide the groups from observed states of agents based on the Frechet distance. 

The group behavior is important information to the trajectory prediction \cite{Pellegrini2020}, which constrains the motion of the agent deviating from group trajectories. However, the groups' classification cannot be directly observed, which hides in ${\mathbf P}_{t}^i$. There are several methods of group classification, such as the learning methods and the distance evaluation methods. In \cite{Yamaguchi2011,Alexandre2016}, the Support Vector Machine classifier, which defines the quantity features as the normal distance, the normal velocity, the normal relative direction, and the time-overlap ratio with the observed pair of trajectories, is used to train the classifier. Based on the unsupervised clustering, the feature of the coherent motion calculated by observed trajectories is employed to divide agents' groups \cite{Dehghan2018}. In \cite{Gaoshan2017}, two agents are classified as the same group when their distances of observed nodes are less than a constant threshold. These methods need lots of data to train or have low accuracy. Limited by the calculation time and the amount of observed data, the Frechet distance is utilized to classify groups from observed states.

The Frechet distance has excellent performance at the measurement of similarity for curves. For the given continuous trajectories $L_1$ and $L_2$, their Frechet distance is defined as
\begin{equation}
{\delta _{\rm{F}}}({L_1},{L_2}) = \mathop {\inf \max }\limits_{\alpha ,\beta ,t_f \in [0,1]} d\left( {{L_1}(\alpha (t_f)),{L_2}(\beta (t_f))} \right),
\label{E16_c_Fdist}
\end{equation}
where $t_f$ is regarded as sampling time, $\alpha(\cdot)$ and $\beta(\cdot)$ are continuous non-decreasing functions which map the $t$ to the points in $L_1$ and $L_2$, and $d(\cdot)$ is the Euclidean distance.

\begin{figure}[htb]
	\centering
	\subfigure[]{
		\includegraphics[width=8cm]{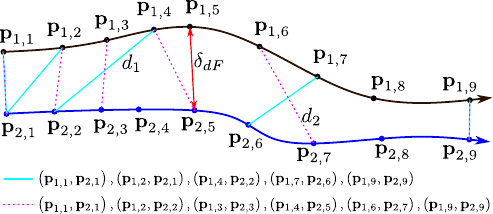} 
		\label{F_new3_FrechetDis_process}
	}
	\quad
	\subfigure[]{
		\includegraphics[width=8cm]{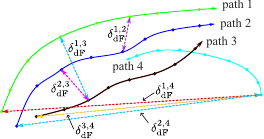}	
		\label{F_new3_21_FrechetDis_similarity}
	}
	\quad
	\subfigure[]{
		\includegraphics[height=4cm]{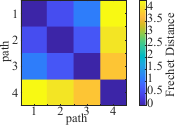}	
		\label{F_new3_22_FrechetDis_similarity}
	}
	\caption{The discrete Frechet distance and the similarity comparing. (a) The process of calculating the discrete Frechet distance between two point sets. The black and blue arrow curves are two paths, and the points on the curves are sampled path nodes. To calculate the discrete Frechet distance, there display two mapping modes, $a_k$ and $b_k$, which are connected by cyan lines and red dashed lines, for example. In the mapping mode with cyan line, ${\rm N}=5$ and the value of \eqref{E17_maxDis} is $d_1$, which is the distance between ${\bf{p}}_{1,4}$ and ${\bf{p}}_{2,2}$. As the mapping mode with red dashed lines, ${\rm N}=6$ and the value of \eqref{E17_maxDis} is ${d_2} = \left\| {{{\bf{p}}_{1,6}} - {{\bf{p}}_{2,7}}} \right\|$. ${\delta _{{\rm{dF}}}}$ is the discrete Frechet distance, which is the distance between ${\bf{p}}_{1,5}$ and ${\bf{p}}_{2,5}$, under the given sampled path nodes. (b) The comparison of paths' similarities by the Frechet distance. The double arrow dashed lines represent values of the Frechet distances corresponding to compared paths. The Frechet distance ${\delta _{{\rm{dF}}}^{1,2}}$ of path 1 and path 2 is the smallest, and the Frechet distances between path 4 to the other three paths are larger than ${\delta _{{\rm{dF}}}^{1,2}}$, ${\delta _{{\rm{dF}}}^{1,3}}$ and ${\delta _{{\rm{dF}}}^{2,3}}$. (c) The similarities of paths with the Frechet distances in (b), the smaller value the higher similarity.}
\end{figure}

However, our observed states are discrete. In this case, their Frechet distances are approximate values defined as
\begin{subequations}
\begin{gather}
{d_l} = \mathop {\max }\limits_{k = 1,2,...{\rm N}} d\left( {{{\bf{P}}_1}({a_k}),{{\bf{P}}_2}({b_k})} \right),\label{E17_maxDis}\\
{\delta _{{\rm{dF}}}}({{\bf{P}}_1},{{\bf{P}}_2}) = \mathop {\min}\limits_l \left( {{d_l}} \right),\label{E18_discrete_F}
\end{gather}
\end{subequations}
where $k$ is the discrete sampling time, ${\bf{P}}_1$ (or ${\bf{P}}_2$) is a set of sampled positions $\{{\bf{p}}_1,{\bf{p}}_2,...,{\bf{p}}_{\rm N}\}$, $a_k$ and $b_k$ are the same meanings as $\alpha$ and $\beta$ in \eqref{E16_c_Fdist}, $d_l$ is the maximum distance under the $ l$th combination of $a_k$ and $b_k$, and ${\delta _{{\rm{dF}}}}(\cdot)$ is the discrete Frechet distance between ${\bf{P}}_1$ and ${\bf{P}}_2$. ${\delta _{{\rm{dF}}}}(\cdot)$ represents the similarity value of two given trajectories. The process of calculating the Frechet distance by \eqref{E17_maxDis} and \eqref{E18_discrete_F} is shown in \mbox{Fig. \ref{F_new3_FrechetDis_process}}.

To acquire the groups' information from observed states, we first use \eqref {E17_maxDis} and \eqref{E18_discrete_F} to calculate the Frechet distance ${\delta _{{\rm{dF}}}^{i,j}}$ of every two observed agents $i$ and $j$, and build the similarity matrix $\bf \Psi $ as
\begin{equation}
{\bf \Psi}  = \left[ {\begin{array}{*{20}{c}}
\vspace{1ex}
{\delta _{{\rm{dF}}}^{1,1}}&{\delta _{{\rm{dF}}}^{1,2}}& \cdots &{\delta _{{\rm{dF}}}^{1,{{\rm{M}}_t}}}\\
{\delta _{{\rm{dF}}}^{2,1}}&{\delta _{{\rm{dF}}}^{2,2}}& \cdots &{\delta _{{\rm{dF}}}^{2,{{\rm{M}}_t}}}\\
 \vdots & \vdots & \ddots & \vdots \\
{\delta _{{\rm{dF}}}^{{{\rm{M}}_t},1}}&{\delta _{{\rm{dF}}}^{{{\rm{M}}_t},2}}& \cdots &{\delta _{{\rm{dF}}}^{{{\rm{M}}_t},{{\rm{M}}_t}}}
\end{array}} \right],
\label{E19_similaritymatrix}
\end{equation}
where ${{\rm{M}}_t}$ is the number of observed agents at time $t$. This process is shown in \mbox{Fig. \ref{F_new3_21_FrechetDis_similarity}} and 3(c). In particular, from \mbox{Fig. \ref{F_new3_22_FrechetDis_similarity}}, it is convenient to compare the differences of similarities in $\bf \Psi $.

Then, a constant threshold value $\Delta {\delta _{{\rm{dF}}}}$ is used to judge whether agents $i$ and $j$ belong to the same group (if ${\delta _{{\rm{dF}}}^{i,j}} \le \Delta {\delta _{{\rm{dF}}}}$, they are in one group). For the efficiency of group classification, we use the binary function to represent \eqref{E19_similaritymatrix} as
\begin{equation}
{\bf \Psi}_0^{i,j} = \left\{ \begin{array}{ll} \vspace{1ex}
1,&{\bf \Psi} ^{i,j}  \le  \Delta {\delta _{{\rm{dF}}}}\,\,{\rm{and}}\,\,i\ne j\\
0,&{\rm{otherwise}}
\end{array} \right..
\label{E20_binaryP}
\end{equation}

Finally, the desired speed values of agents in \eqref{E3_speedfunction} are estimated by observed states as
\begin{equation}
{u_i} = \sum\limits_{k = -{\rm N}+2}^{0} {{{W}_{k+{\rm N}-1}} \cdot \left\| {\frac{{{\bf{p}}_{t+k}^i - {\bf{p}}_{t+k-1}^i}}{{\Delta t}}} \right\|},
\label{E22_desiredV}
\end{equation}
where $W_{k+{\rm N}-1}>0$ is the weight parameter, $\sum\limits{W_{k+{\rm N}-1}}=1$. Based on \eqref{E7new_calculatedGp}, the average speed value of the $l$th group is
\begin{equation}
{u_l} = \frac{1}{{{m_l}}}\sum\limits_{i \in G_l} {{u_i}},
\label{E21_groupV}
\end{equation}
where $m_l$ indicates the number of agents in the $l$th group, and $G_l$ is the agents' indices set of agents in $l$th group.

\begin{algorithm}[htb]
\label{Alg1_group}
\caption{\texttt{Group\_information}}
\KwIn{Observed positions $\{{\bf{P}}_t^i\}$, and number of agents ${\rm{M}}_t$}
\KwOut{Group classification $\{G_l\}$ and group average speed $\{u_l\}$}
Calculate the Frechet distances by \eqref {E17_maxDis} and \eqref{E18_discrete_F} under $\{{\bf{P}}_t^i\}$\;

Build the similarity matrix ${\bf \Psi}$ by \eqref{E19_similaritymatrix}\;

Calculate the binary matrix ${\bf \Psi} _0$ of $\Psi$ by \eqref{E20_binaryP}\;

Agents are numbered from 1 to ${\rm{M}}_t$\;

Initialize the active vector ${{\bf F}_{ active}}=[1,2,...,{\rm{M}}_t]$ and group number ${\rm N}_{ group}=0$\;

\While{${\bf F}_{active}$ is not empty}
{
Select the first element in ${{\bf F}_{active}}$ as the basic index ${\rm{No}}_a$ \;

Find in the ${\rm{No}}_s$th row of ${\bf \Psi} _0$ the indices of the elements with value 1, and put those indices in a vector ${{\bf F}_{basic}}$. Call ${\rm{N}}_f$ the length of ${{\bf F}_{basic}}$\;

\If {${\rm{N}}_f \ge 1$}
{
Set the current new group number as ${\rm N}_{group}={\rm N}_{group}+1$\;

Store elements in the ${\rm N}_{group}$th group as ${{\bf F}_{En}}=[{\rm{No}}_a,{{\bf F}_{basic}}]$\;

\While{${{\bf F}_{basic}}$ is not empty}
{
Select the first element in ${{\bf F}_{basic}}$ as ${\rm{No}}_s$, and remove it from ${{\bf F}_{basic}}$ simultaneously\;

Find in the ${\rm{No}}_s$th row of ${\bf \Psi} _0$ the indices of the elements with value 1, and put those indices in a vector ${{\bf F}_{same}}$. Call ${\rm{N}}_s$ the length of ${{\bf F}_{same}}$\;

\For{$r=[1,...,{\rm{N}}_s]$ \rm{and} ${\rm{N}}_s \ne 0$}
{
	\If{${{\bf F}_{same}}(r) $ is not an element in $ {{\bf F}_{En}}$}
	{
		Add the index ${{\bf F}_{same}}(r)$ to ${{\bf F}_{basic}}$ and ${{\bf F}_{En}}$\;
	}
} 
} 
Add the group ${{\bf F}_{En}}$ and number $l = {\rm N}_{group}$ to ${{G}}_l$\;
Remove elements in ${{\bf F}_{En}}$ from ${{\bf F}_{active}}$\;
} 
\Else
{
Remove the first element in ${{\bf F}_{active}}$\;
}
}  

\For{$l=[1,...,{\rm N}_{group}]$ \rm{and} ${\rm N}_{group} > 0$}
{
Calculate the average speed value $u_l$ of the $l$th group by \eqref{E21_groupV}\;
}
\end{algorithm}
The details of groups' information estimation are shown in \textbf{Algorithm} \ref{Alg1_group}, where the roles of utilized vectors are listed as follows:
\begin{itemize}
\item ${\bf F}_{active}$ is a vector that lists the agents in numerical order.
\item ${\bf F}_{basic}$ is a vector that stores the raw indices that have not been searched in ${\bf F}_{active}$ to check for other indices.
\item ${\bf F}_{En}$ is a vector that stores the indices belonging to the current group.
\item ${\bf F}_{same}$ is a vector that stores the searched indices based on ${\rm No}_s$.
\end{itemize}

In \textbf{Algorithm} \ref{Alg1_group}, we firstly calculate the similarity matrix (Line 1 - Line 3), and  initialize the active vector (Line 4 - Line 5). Then agents are classified (Line 6 - Line 21), and the inner loop (Line 12 - Line 19) searches for agents in the same group. Finally, the loop (Line 22 - Line 23) calculates the average speeds of groups.

\begin{figure}[htb]
	\centering
	\includegraphics[width=8.8cm]{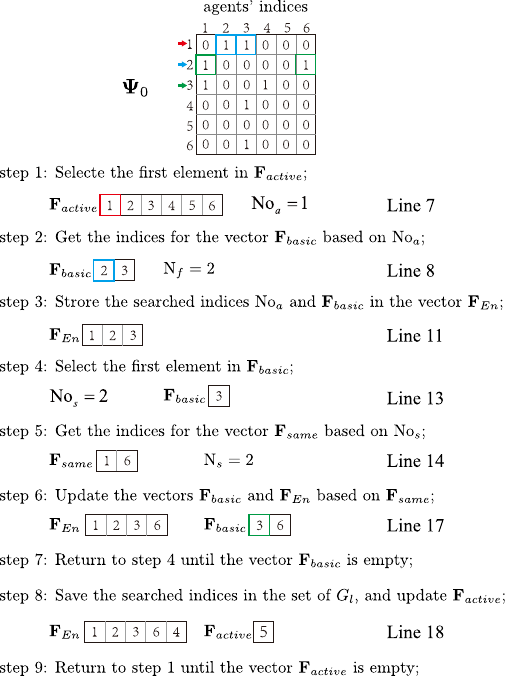}	 
	\caption{The workflow of \textbf{Algorithm} \ref{Alg1_group}.}
	\label{F4_n_process_alg1}
\end{figure}

To clearly explain the group division by the Frechet distance, \mbox{Fig. \ref{F4_n_process_alg1}} displays an example that illustrates the process of \textbf{Algorithm} \ref{Alg1_group}. Based on the binary matrix ${\bf \Psi} _0$ shown on the top of the figure and the active vector ${{\bf F}_{active}}$, firstly, we set ${\rm{No}}_a=1$, and get the vector ${{\bf F}_{basic}}=[2,3]$ from the first column in ${\bf \Psi} _0$ by Line 8. Secondly, the current elements in the group are recorded in ${{\bf F}_{En}}=[1,2,3]$. Thirdly, we get the first element ${\rm{No}}_s=2$ in ${{\bf F}_{basic}}$, and remove it from ${{\bf F}_{basic}}$. Then, in the ${\rm{No}}_s$th row of ${\bf \Psi} _0$, the agents' indices in the same group are ${{\bf F}_{same}}=[1,6]$, where ${{\bf F}_{same}}$ stores column indices whose values equal to $1$ in the row index of ${\rm{No}}_s$ (Lines 13-14). Next, to prevent redundant searching, we select the elements of ${{\bf F}_{same}}$ that have not been checked in ${{\bf F}_{En}}$ (Lines 15-16). Then we add the index ${{\bf F}_{same}}(r)$ that satisfies the condition in Line 16 to ${{\bf F}_{En}}$ and ${{\bf F}_{basic}}$, where ${{\bf F}_{En}}$ stores the agents' index that in the current group, ${{\bf F}_{basic}}$ stores the row indices that have not been searched to check whether there are other indices belonging to the current group (Line 17). Execute the loop (Lines 12-19) until the vector ${{\bf F}_{basic}}$ is empty; this means the agents in the one group ${{\bf F}_{En}}=[1,2,3,6,4]$ have been found. Finally, check the remaining elements in ${{\bf F}_{active}}$ to find other groups until ${{\bf F}_{active}}$ is empty.

\section{Parameters Estimation}\label{Variabletrain}
This section decouples the optimization problem to calculate the optimal velocities and the parameters separately.

\subsection{Optimization problem decoupling strategy}
In this study, one of the most critical problem is to estimate the approximate optimal parameters in the sets $\Theta ^*$ in \eqref{E15_trainpara}. Although the group information has been estimated, the information of agents' future destinations after the current time $t$ is still unknown. To estimate parameters in the sets $\Theta$, we need to optimize \eqref{F14_optV} for all agent under each optimal set $\Theta_i^*$. However, it will cost lots of time to estimate $\Theta _i^*$, which prevents real-time trajectory prediction.

The function \eqref{E15_trainpara} is too complex to be exactly optimized. For different values in $\Theta_i$, one needs to recalculate some values (i.e., ${{\bf{z}}_i} - {{\bf{p}}_{t+k-1}^i}$, $\Delta {\bf{p}}_{t+k-1}^{i,j}$) from \eqref{E2_dampfunction} to \eqref{E6_groupfunction} and in \eqref{E10_Einterdw}, which is rather inefficient. To simplify \eqref{E15_trainpara}, we define the optimal ${\bf{v}}_{t + k}^{{i_*}}$ as:
\begin{equation}
{\bf{v}}_{t + k}^{{i_*}} = {{{\bf{v}}_{t + k}^{{i_*}}}({\bf p}_{t + k - 1}^i,\{ {\bf p}_{t + k - 1}^j\} ,{\Theta _i})}.
\label{E22_newV}
\end{equation}
In \eqref{E15_trainpara}, the optimal predicted velocity at the time $t+ k$ is calculated with predicted state $\bar {\bf p}_{t + k - 1}^i$. Here, we utilize the real state ${\bf p}_{t + k - 1}^i$ to replace the predicted state. This will make the value of \eqref{E12_newEnergy} only rely on ${\bf{v}}_{t + k}^{{i_*}}$ no matter what the predicted state $\bar {\bf p}_{t+ k - 1}^i$ is, and the values, such as ${{\bf{z}}_i} - {{\bf{p}}_{t + k - 1}^i}$, $\Delta {\bf{p}}_{t + k - 1}^{i,j}$ and ${\rm{D}}(\Delta {\bf{p}}_{t + k - 1}^{{{ij}}})$, are calculated just one time for every agent. By combining \eqref{E22_newV}, the set $\Theta_i$ in \eqref{E15_trainpara} is estimated as follows:
\begin{equation}
\Theta _i^* = \mathop {\arg \min }\limits_{{\Theta _i}} {\sum\limits_{k = -{\rm N} +2} {\left\| {{\bf{v}}_{t+ k}^i - {\bf{v}}_{t + k}^{{i_*}}} \right\|} ^2},
\label{E23_newOpt}
\end{equation}
where ${{\bf{v}}_{t + k}^i}$ is the real velocity calculated by observed states. There is another difference between \eqref{E15_trainpara} and \eqref{E23_newOpt}: the former needs to calculate each ${\bf{\bar p}}_{t+ k}^i$ one by one for the sampling trajectory of agent $i$, but the latter enables the parallel computation of ${{\bf{v}}_{t+ k}^{{i_*}}}$ during the time range $t-{\rm N} +2$ to $t$.

When calculating the value of \eqref{E12_newEnergy}, the sub-energy functions that need large quantities of multiple operations are \eqref{E5_attractfunction} and \eqref{E7_interactionfunction}, in which the times of sum operations are ${\rm{N}_i^g}+({\rm{M}}_t+{\rm{N}_{obst}})$. Here, ${\rm{N}_i^g}$ is the number of agents which stay in the same group with agent $i$, and ${\rm{N}_{obst}}$ is the number of static environment obstacles shown in \mbox{Fig. \ref{F3_obst_headselect}}. Based on the iteration operation in \textbf{Algorithm} \ref{Alg2_theta_opt} and \textbf{Algorithm} \ref{Alg3_V_opt}, the total computation times of \eqref{E5_attractfunction} and \eqref{E7_interactionfunction} are $({\rm{N}_i^g}+({\rm{M}}_t+{\rm{N}_{obst}}))$ when using the improved model \eqref{E22_newV} to train the parameters for each agent. As for the parameters training scheme by \eqref{E15_trainpara}, there would be ${\rm L}\cdot{\rm N}_{\Theta}\cdot{({\rm N}-1)}\cdot({\rm{N}_i^g}+({\rm{M}}_t+{\rm{N}_{obst}}))$ times to calculate \eqref{E5_attractfunction} and \eqref{E7_interactionfunction}. According to the above analysis, it can be seen that the function \eqref{E22_newV} can save a large number of computing resources, and the side effect is a little bit reduction of computation accuracy, which is acceptable according to the experiment results.

Next, before estimating the parameters in $\Theta _i^*$, it is necessary to deal with the unknown destination ${\bf{z}}_i$. Note that agents' motion characteristics in \eqref{E12_newEnergy} are not unique \cite{Alexandre2016}; the parameters in $\Theta _i^*$ change with the environment and the agents' thoughts, etc. In \cite{Pang2021}, the prior states, which are resampled in observed historical states, are utilized to train the prediction model. Without loss of generality, we replace the future destination ${\bf{z}}_i$ with the position in the last state of the current observed ${\bf P}_{t}^i$ to train $\Theta _i$. The reason is that the last observed state ${\bf p}_t^i$ can be considered as a sub-target position of agent $i$ at the start of observing time $t-{\rm N}+1$ with the global view. In this case, we can estimate the approximate optimal parameters in $\Theta _i^*$ of agent $i$ over a period of time.

\subsection{SSA-based optimization method}
Based on the above analysis, the estimation of optimal parameters in $\Theta _i$ comprises two optimizing loops. The outer loop in \eqref{E23_newOpt} and inner loop in \eqref{F14_optV} are utilized to evaluate the cost of velocities' errors and the value of \eqref{E12_newEnergy}, respectively.

The outer loop is a high-dimension and nonlinear optimization problem, for there are eight elements in $\Theta _i$. Traditional methods, such as gradient descent, are hard to solve the problem. Meta-heuristic algorithms, which include Ant Colony Optimization (ACO) and Particle Swarm Optimization (PSO), are powerful methods to solve this problem. However, ACO and PSO are easy to fall into the local optima and have multiple control parameters. 

In this paper, the salp swarm algorithm is employed to optimize $\Theta _i$. SSA is a bio-inspired optimizer that simulates the process of salp-swarm predation to solve complex optimization problems \cite{Mirjalili2017}. The initial states of the swarm are defined as ${{\bf{X}}_0} = \left[ {{\bf{x}}_0^1,{\bf{x}}_0^2,...,{\bf{x}}_0^{{\rm N}_{\Theta}}} \right]$; ${\rm{N}}_{\Theta}$ is the number of salps in the swarm; ${\bf{x}}_0^1$ is the leader's state, and the others are followers' states. Driven by the food, their states are updated by:
\begin{subequations}
\begin{gather}
{\bf{x}}_{\rm{k}}^1 \!=\! {\bf{x}}_{{\rm{k}} - 1}^{\rm{F}} \!+\! {\sign {\bf{c}_3}}\cdot {{\rm{c}}_1} \!\cdot\!\left( {({{\bf{x}}_{\max }} \!-\! {{\bf{x}}_{\min }}) \cdot {{\bf{c}}_2} \!+\! {{\bf{x}}_{\min }}} \right),\label{E24_Lsalpupdate}\\
{\bf{x}}_{\rm{k}}^i = \frac{1}{2}({\bf{x}}_{{\rm{k -1}}}^i + {\bf{x}}_{\rm{k}}^{i - 1}),i=[2,3...,{\rm{N}}_{\Theta}],\label{E24_Fsalpupdate}
\end{gather}
\end{subequations}
with $\rm{k}$ being the iteration time, ${\bf{c}_2}$ and ${\bf{c}_3}$ are random vectors generated by uniform distributions $[0,1]$ and [-1,1], ${\bf{x}}_{{\rm{k}} - 1}^{\rm{F}}$ is the state of the food, ${\bf{x}}_{\max }$ and ${\bf{x}}_{\min }$ are bounds of salps' states, and ${\rm{c}}_1$ is a weight parameter defined as:
\begin{equation}
{{\rm{c}}_1} = 2\cdot \exp {\left( { - \frac{{{\rm{4k}}}}{{\rm{L}}}} \right)^2},
\label{E25_c1}
\end{equation}
where $\rm{L}$ is the maximum iteration time. From \eqref{E24_Fsalpupdate}, we can see that the followers move with their front one. When the individual is in the back state of ${\bf{x}}_{\rm{k}}^i$, its position change will be slow, making the swarm can explore and expand in the search space.

\begin{algorithm}[htb]
\label{Alg2_theta_opt}
\caption{\texttt{Opt\_parameters\_estimation}}
\KwIn{Observed states $\{{\mathbf P}_{t}^i\}$, group information $\{G_l\}$ and $\{u_l\}$}
\KwOut{Approximate optimal parameters in $\{\Theta_i^*\}$}
\For{$i=[1,2,...,{\rm{M}}_t]$}
{
Based on ${\mathbf P}_{t}^i$, ${\{{\mathbf  P}_{t}^j\}_{j \ne i}}$, $\{G_l\}$ and $\{u_l\}$, calculating the static values $\{{\bf A}_k\}$ such as ${{\bf{z}}_i} - {{\bf{p}}_{t+k-1}^i}$, $\Delta {\bf{p}}_{t+k-1}^{i,j}$ and ${\rm{D}}(\Delta {\bf{p}}_{t+k-1}^{{{ij}}})$ in \eqref{E12_newEnergy}\;

Set the destination ${\bf{z}}_i$ as ${\bf{p}}_t^i$\;

Initialize ${\bf{\Theta}}_i = [\Theta_i^1,\Theta_i^2,...,\Theta_i^{{\rm{N}}_{\Theta}}]$ randomly under bounds $\Theta_{\max}$ and $\Theta_{\min}$, best cost ${\rm{C}}_{best}$, optimal parameters in $\Theta_i^*$\;

\For{${\rm{k}} = [1,2,...,\rm{L}]$}
{
\For{$n = [1,2,...,{\rm{N}}_{\Theta}]$}
{
${{\bf{V}}_i^n} = \{\texttt{Opt\_velocity\_estimation}(\cdot)\}$\;
Evaluate ${{\bf{V}}_i^n}$ by \eqref{E23_newOpt}, and put the cost value to ${\bf{C}}_{array}$\;
}  
Select the minimum value ${\rm{C}}_m$ in ${\bf{C}}_{array}$ and its corresponding set $\Theta_i^m$ in ${\bf{\Theta}}_i $\;
\If{${\rm{C}}_m \leq {\rm{C}}_{best}$}
{
$\Theta_i^* = \Theta_i^m$\;
${\rm{C}}_{best} = {\rm{C}}_m$\;
}  
Calculate ${{\rm{c}}_1}$ based on \eqref{E25_c1}\;
Update ${\bf{\Theta}}_i $ by \eqref{E24_Lsalpupdate} and \eqref{E24_Fsalpupdate}\;
} 
}  
\end{algorithm}

The outline of approximate parameters estimation is shown in \textbf{Algorithm} \ref{Alg2_theta_opt}. Lines 2-3 prepare the static values in \eqref{E12_newEnergy}. Line 4 initializes ${\rm{N}}_{\Theta}$ sets of $\Theta_i$. Lines 6-8 sample ${{\bf{V}}_i^n} = [{\bf{v}}_{t-{\rm N} + 2}^{{i_*}},...,{\bf{v}}_t^{{i_*}}]$ based on each $\Theta_i^n$ and evaluate their costs with real sampled velocities. Lines 9-14 select the optimal set $\Theta_i^m$ as the food to update swarm's positions.

\section{Velocity Estimation}\label{S_OptSpeedEst}
In this section, we design the inner loop optimization problem in \eqref{F14_optV} by the GD-based SSA methods to estimate the optimal velocities under the given set $\Theta_i$.

The inner loop is a nonlinear function with two variables ${\bf{v}} = [v_x,v_y]^{\rm T}$. In \cite{Yamaguchi2011,Alexandre2016}, the gradient descent is used to find the optimal value in \eqref{F14_optV}. However, it may fall into the local optima and converge slowly. To deal with the problems, we combine SSA with GD to design a new optimization method that fuses their abilities of extensive exploration and reliable convergence. In SSA, the optimal state ${\bf{x}}_{{\rm{k}}}^{\rm{F}}$ is acquired from the sampled states ${{\bf{X}}_{{\rm{k}} - 1}}$. Based on this step, we implement the GD method beginning with ${\bf{x}}_{{\rm{k}}}^{\rm{F}}$ to find the optimal state. 

\begin{algorithm}[htb]
\label{Alg3_V_opt}
\caption{\texttt{Opt\_velocity\_estimation}}
\KwIn{Last velocity ${\bf{v}}_{t+k-1}^i$, sampled parameter $\Theta_i$, pre-calculated values ${\bf A}_{k}$}
\KwOut{Approximate optimal velocity ${\bf{v}}_{t+k}^{i_*}$}

Initialize the max iteration step ${\rm L}_{\rm v}$, the number of salps in the swarm ${\rm{N}}_{\rm v}$\;

Initialize ${\bf V}_{2:{{\rm{N}}_{\rm v}}} = [{\bf v}_2,...,{\bf v}_{{\rm{N}}_{\rm v}}]$ randomly under bounds ${\bf v}_{\max}$ and ${\bf v}_{\min}$, ${\bf V}_1 = {\bf{v}}_{t+k-1}^i$ and best cost ${\rm C}_{best}^{\rm v}$\;

\For{${\rm k} = [1,2,...,{\rm{L}}_{\rm v}]$}
{
Evaluate ${\bf V}$ by \eqref{E12_newEnergy}, and put cost values to ${\bf C}_{array}^{\rm v}$\;
Select the minimum value ${\rm C}_m^{\rm v}$, and its corresponding sample ${\bf v}_m$\;
\If{${\rm C}_m^{\rm v} \leq {\rm C}_{best}^{\rm v}$}
{
${\bf{v}}_{t+k}^{i_*} = {\texttt gradient\_descent}({\bf v}_m)$\;
Calculate the cost value ${\rm C}_{best}^{\rm v}$ of ${\bf{v}}_{t+k}^{i_*}$\;
}
Calculate ${{\rm{c}}_1}$ based on \eqref{E25_c1}\;
Update ${\bf V}$ by \eqref{E24_Lsalpupdate} and \eqref{E24_Fsalpupdate} with ${\bf{v}}_{t+k}^{i_*}$\;
}
\end{algorithm}

The outline of approximate optimal velocity estimation is shown in \textbf{Algorithm} \ref{Alg3_V_opt}. Lines 1-2 initialize the parameters and sample states $\bf V$ of the swarm. Particularly, the first element ${\bf v}_1$ in ${\bf V}$ is initialized by ${\bf{v}}_{t+k-1}^i$, which can enhance the convergence rate of estimation, for the agents do not move drastically in general. Lines 3-10 try to search for the approximate optimal velocity ${\bf{v}}_{t+k}^{i_*}$. Lines 6-8 utilize the GD method to find the optimal velocity, which begins with the sampled ${\bf v}_m$.

\section{Target Headings Estimation}\label{Sec4_goal_traj}
This section first presents the goal style we have proposed and the way of estimating agents' goals. Then we present the whole estimation method of agents' future trajectory prediction.

\subsection{Goal type and estimation}\label{Goal_estimation}

In Section \ref{Variabletrain}, the last state in ${\mathbf P}_{t}^i $ is defined as the destination; however, it cannot be utilized to predict future states, since it represents the desired position in the previous observation period rather than the future target position starting from the current time. In most studies, the goal type used to predict agents' future states is the position in space coordinate \cite{David2020, Liang2020}. In this case, the Bayesian inference \cite{Bashar2018} and learning-based methods \cite{Zhitian2021} are employed to estimate agents' future target positions. Nevertheless, it is hard to determine agents' reliable destinations with a small amount of data in complex environments. 

Different from existing studies, we select the target headings as the agents' goal. It is determined under the assumption that agents always move with the constant target headings until they want to change their goal destinations. Although the whole trajectories of some agents are not straight lines, for agents can go anywhere they want, we only focus on the target headings within a period. In particular, with the assumption, it enables trajectory prediction based on the target heading. Compared with the final destination estimation, the target heading estimation has one less dimension (in the two-dimensional Cartesian coordinate system, the target heading belongs to one dimension space) and more tolerance to the agents' uncertain states.

\begin{figure}[htb]
	\centering
   \def\svgwidth{8.4cm}
	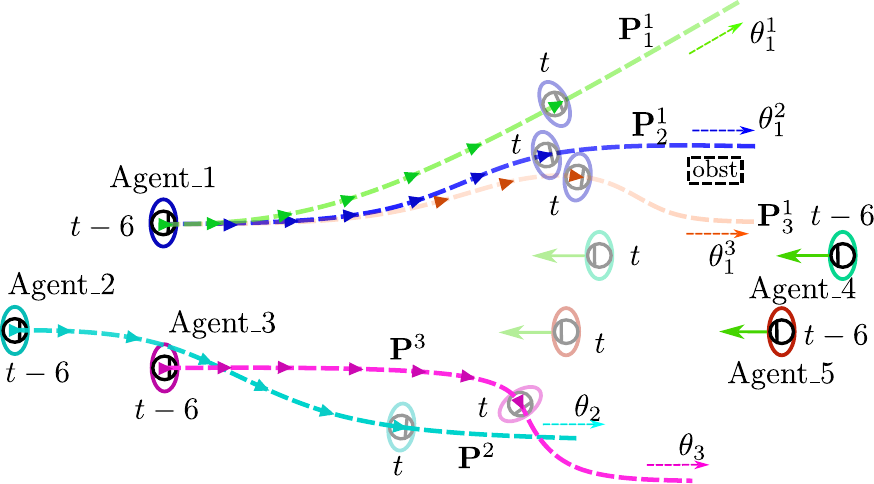
	\caption{Interactive motions of agents. The curved lines with dashed are trajectories of agents, the triangles represent the sampled nodes in trajectories, the arrow dotted lines refer to agents' target headings, the cyan arrows are moving directions of agents 4 and 5, and the black dashed box represents the uncertain static obstacle to the trajectory ${\bf P}_3^1$.}
	\label{F2_agentinter_short}
\end{figure}

To estimate agents' target headings, one of the simplest methods is to calculate the agent's current heading as the goal. Another more effective method is to use the whole observed states to compute the average heading. Obviously, the two methods produce enormous prediction errors with high probabilities, which can be seen from \mbox{Fig. \ref{F2_agentinter_short}}. For example, when agent 3 interacts with agent 5 at $t$, the current heading of agent 3 is far away from its target heading $\theta _3$; by using the aforementioned two methods, the current heading will affect the calculated target heading too significantly.

In this paper, we introduce a new method named similarity-based resample evaluation (SRE) to estimate agents' target headings. It is known that the future information hides in observed states, so with the estimated sets $\Theta^*$ in Section \ref{Sec3_group}, we can replace the goal type in \eqref{E4_destinationfunction} and resample agents' paths from $t-{\rm N}+2$ to $t$. Note that the term $\frac{({{\bf{z}}_i} - {\bf{p}}_{t + k}^i)^{\rm T}}{{\|{{\bf{z}}_i} - {\bf{p}}_{t + k}^i\|}}$, which is the first term in the right-hand side of equation \eqref{E4_destinationfunction}, is the normalized direction vector from the current position to the destination. This means that replacing $\frac{({{\bf{z}}_i} - {\bf{p}}_{t + k}^i)^{\rm T}}{{\|{{\bf{z}}_i} - {\bf{p}}_{t + k}^i\|}}$ by the term $\left[ {\cos (\theta _i^g),\sin (\theta _i^g)} \right]$ has similar effects on the cost of ${{\rm{E}}_{direction}}$. To estimate the target heading, we sample a heading $\theta_i^g$ from $[-\pi,\pi]$ and then redefine \eqref{E4_destinationfunction} as
\begin{equation}
{{\rm{E'}}_{direction}}({\bf{v}}_{t +k+ 1}^i) =  - \left[ {\cos (\theta _i^g),\sin (\theta _i^g)} \right] \cdot \frac{{{\bf{v}}_{t +k+ 1}^i}}{{\|{\bf{v}}_{t +k+ 1}^i\|}}.
\label{E25_headingE2}
\end{equation}
It shows that ${{\rm{E'}}_{direction}}({\bf{v}}_{t+k + 1}^i)$ is of low cost when the direction of ${\bf{v}}_{t+k + 1}^i$ is close to $\theta _i^g$. Then, the energy function can be written as
\begin{equation}
\begin{aligned} 
{{\rm{E''}}_{{\Theta}_i^*} }(\cdot) =& {\lambda _0^i}{{\rm{E}}_{damp}}(\cdot)+ 
{\lambda _1^i}{{\rm{E}}_{speed}}(\cdot) + \\
&{\lambda _2^i}{{\rm{E'}}_{direction}}(\cdot)+{\lambda _3^i}{{\rm{E}}_{attraction}}(\cdot)+ \\
&{\lambda _4^i}{{\rm{E}}_{group}}(\cdot)+ 
{{\rm{E'}}_{collision}(\cdot)}.
\end{aligned} 
\label{E26_headenergycost}
\end{equation}
In \eqref{E26_headenergycost}, the parameters of the energy function are the trained ${\Theta}_i^*$. The sampled path ${{\bf{\tilde P}}^i} = [{\bf{\tilde p}}_{t -{\rm N}+2}^i,{\bf{\tilde p}}_{t -{\rm N}+ 3}^i,...,{\bf{\tilde p}}_t^i]$ of agent $i$ is based on ${\Theta}_i^*$ and $\theta _i^g$ starts at $t -{\rm N}+2$, and \eqref{F14_optV} updates the positions of agent $i$ until $t$. In this process, the states of other agents interacting with agent $i$ are the ground truths. 

Next, we need to quantify the cost caused by $\theta _i^g$ based on ${{\bf{\tilde P}}^i}$ and find the target heading $\theta _i^{g_*}$. Due to the target heading hiding in the sampled path ${{\bf{\tilde P}}^i}$, it is natural that the sampled $\theta _i^g$ can be evaluated by the similarity between ${{\bf{\tilde P}}^i}$ and ${\bf{P}}_{t}^i$.
Section \ref{Sec3_group} has introduced the Frechet distance to calculate paths' similarity values; however, the group division focuses on the real paths, which is different from the evaluation process between the sampled path and the real path of agent $i$. When ${{\bf{\tilde P}}^i}$ is calculated under approximate optimal set ${\Theta}_i^*$, if the sampled target headings are near the real target heading $\theta _i^{g_*}$, it will make the predicted trajectory not far from the real path ${\bf{P}}_{t}^i$. In this case, we evaluate $\theta _i^g$ by the Frechet distance and Euclidean distance. The cost function is defined as
\begin{equation}
{\rm{C}}_i^H = \eta  \cdot {\delta _{{\rm{dF}}}}\left( {{{\bf{P}}_{t}^i},{\bf{\tilde P}}^i} \right) + (1 - \eta ) \cdot \sum\limits_{k = -{\rm N}+2} {d\left( {{\bf{p}}_{t+k}^i,{\bf{\tilde p}}_{t+k}^i} \right)} ,
\label{E27_costTarHead}
\end{equation}
where $\eta \in [0,1]$ is the weight parameter, and $d\left(\cdot\right)$ is the Euclidean distance between two points. In particular, the optimal target heading is:
\begin{equation}
\theta _i^{{g_*}} = \mathop {\arg \min }\limits_{\theta _i^g} {\rm{C}}_i^H\left( {\theta _i^g|{{\bf{P}}_{t}^i},{\Theta}_i^*} \right).
\label{E28_opttarHead}
\end{equation}

Based on the above analysis, the estimation of the target heading $\theta _i^{{g_*}}$ by the SRE method has fully utilized the agents' motion behaviors and interaction information. \mbox{Fig. \ref{F2_agentinter_short}} shows relations between observed states and agents' target headings. Firstly, the different trajectories of agent 1 represent the influences on target heading decision under the given interaction characteristic. If agent 1 moves with the true path ${\bf P}_1^1$, we can see that its target heading is $\theta_1^1$ from global view. In this case, with the given set ${\Theta}_1^*$ of agent 1, the cost value ${{\rm{E'}}_{coll}}$ affected by interaction is a small part of \eqref{E26_headenergycost}, and the part of self-influence plays a major role in the velocity decision. By the SRE method, the estimated target heading $\theta _1^{{g_*}}$, which is near $\theta_1^1$, has a lower cost value of ${\rm{C}}_1^H(\cdot)$. For the true path ${\bf P}_2^1$ in other cases, since agent 1 interacts with agent 4 and agent 5, the estimated target heading $\theta _1^{{g_*}}$ will be closed to $\theta_1^2$. In \cite{Alexandre2016}, the agents' paths interacting with others are similar to ${\bf P}_3^1$. However, according to the above assumption, it is generally caused by a hidden obstacle, such as a narrow passage. Secondly, the different interaction sensitivities of agents 2 and 3 illustrate the influences on agents' actions decision under the same target headings. In the interaction processes of agents 2 and 3, agent 2 is more sensitive to agents 4 and 5 than that of agent 3, and there is ${\rm d}^2>{\rm d}^3$ for the interaction distance parameters in \eqref{E12_newvariable}. However, the influenced strengths of agents 2 and 3 are ${\rm w}^2<{\rm w}^3$ in \eqref{E12_newvariable}, which make agent 2 closer to agents 4 and 5 than agent 3 at the stable motion periods after $t$. In this process, the estimated target headings of agents 2 and 3 are in the same direction by the proposed SRE method with different sets $\Theta^*$ of the energy function, which indicates its adaption to complex interactions.

\begin{figure}[htb]
	\centering
   \def\svgwidth{6cm}
	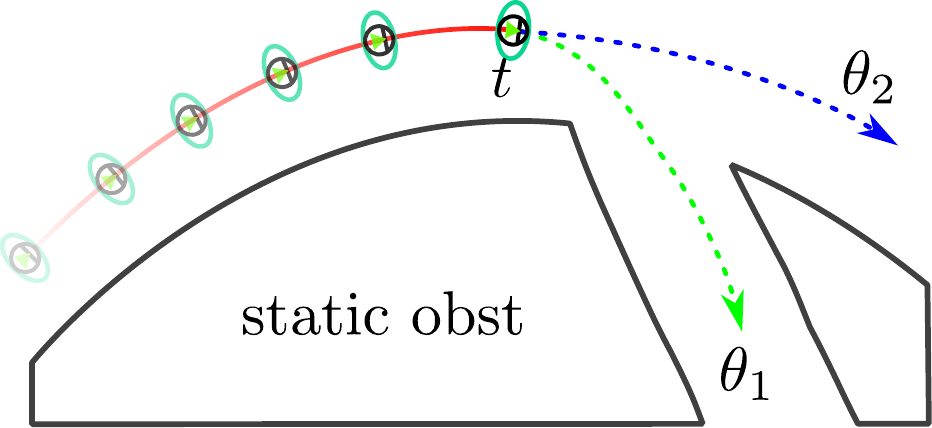
	\caption{Interactive motion between the agent and the environment. The red line is the agent's historical trajectory, the arrow dashed lines represent the possible future trajectories, $\theta_1$ and $\theta_2$ are target headings, the yellow points on the black curves are sampled static environment obstacles.}
	\label{F3_obst_headselect}
\end{figure}

The agent not only interacts with other agents but also with the environment. For example, in \mbox{Fig. \ref{F3_obst_headselect}}, the agent moves along the edge of the obstacle. Based on observed states, the directions $\theta_1$ and $\theta_2$ have high probabilities of being the agent's target heading with the estimation by the SRE method. Therefore, it is valuable to consider the interaction between agents and environment. The interaction can constrain agents' unreasonable actions, such as interaction occurring in forbidden areas,
to improve the reliability of the SRE method. However, the continuous space of obstacle boundaries will affect the velocity of solving \eqref{E28_opttarHead}. Following \cite{Alexandre2016}, we sample positions near the obstacle boundaries and model them as static agents.

Note that it is generally impossible to solve the target heading estimation problem in the continuous space $[-\pi,\pi]$. We need to discretize the heading space and then evaluate the sampled headings to find the approximate optimal target heading. Based on the above assumption, the sampled headings in the opposite direction of the current moving direction are rare probabilities of being the target heading. In this case, it is inefficient to find the target heading in the whole range space. To improve the computation velocity of estimating optimal target headings, we first calculate the average moving direction by:
\begin{equation}
\theta _i^a = \frac{1}{{{\rm N} - 1}}\sum\limits_{k =-{\rm N}+ 2}^{0} {\arctan \frac{{\left\| {x_{t+ k}^i - x_{t+ k-1}^i} \right\|}}{{\left\| {y_{t+ k}^i - y_{t+ k-1}^i} \right\|}}},
\label{E29_averageHead}
\end{equation}
where ${\bf p}_{t+ k}^i ={[{x_{t+ k}^i},{y_{t+ k}^i}]}^{\rm T} \in {\bf P}_{t}^i$. Then, we sample ${\rm N}_{\theta}$ of $\theta_i^g$ on the left and right hand sides of $\theta _i^a$ with a constant angle interval $\Delta{\theta}$. Finally, the sampled headings are utilized to estimate the approximate optimal target heading.

\begin{algorithm}[htb]
\label{Alg4_Target_Head}
\caption{\texttt{Target\_heading\_estimation}}
\KwIn{Optimal set $\Theta_i^*$, observed state ${\mathbf P}_{t}^i$}
\KwOut{Approximate optimal target heading ${\theta}_{i}^{g_*}$}

Sample ${\rm N}_{\theta}$ number of $\theta_i^g$ with \eqref{E29_averageHead}\;

Initial the best cost ${\rm C}_{best}^{\theta}$\;

\For{$\theta_i^{g,n}=[\theta_i^{g,1},\theta_i^{g,2},...,\theta_i^{g,{\rm N}_{\theta}}]$}
{
Initial the current resampled position ${\bf{\tilde p}}_{t-{\rm N}+1}^i = {\bf{p}}_{t-{\rm N}+1}^i$, and the position set ${\bf {\tilde P}}^i$\;
Determine the energy function \eqref{E26_headenergycost} with $\theta_i^{g,n}$ and \eqref{E25_headingE2}\;
\For{$k=[-{\rm N}+2,-{\rm N}+3,...,0]$}
{
${\bf{\tilde v}}_{t+k}^{{i_*}} = {\texttt{Opt\_velocity\_estimation}}(\cdot)$\;

Update ${\bf{\tilde p}}_{t+k}^i$ with ${\bf{\tilde v}}_{t+k}^{{i_*}}$ and \eqref{F13_posupdate}\;

Add the ${\bf{\tilde p}}_{t+k}^i$ to ${\bf {\tilde P}}^i$\;
}
Calculate the cost ${\rm C}_i^H$ of $\theta_i^{g,n}$ by ${\bf {\tilde P}}^i$ and \eqref{E27_costTarHead}\;
\If{${\rm C}_i^H \le {\rm C}_{best}^{\theta}$}
{
${\theta}_{i}^{g_*} = \theta_i^{g,n}$
}
}
\end{algorithm}

The SRE method is displayed in \textbf{Algorithm} \ref{Alg4_Target_Head}. Lines 1-2 show the angle discretization of sampling headings and initialization of the optimal cost value of the sampled target heading. In Lines 4-12, we first initialize the position of the sampled trajectory and build the energy function \eqref{E26_headenergycost} (Lines 4-5). Then, the function ${\texttt{Opt\_velocity\_estimation}}(\cdot)$ in \textbf{Algorithm} \ref{Alg3_V_opt} estimates the approximate optimal velocity with the truth states of other agents and the environment, and the equation \eqref{F13_posupdate} updates the agent's position (Lines 7-9). Finally, we evaluate the sampled heading $\theta_i^{g,n}$ by \eqref{E27_costTarHead} and find the approximate optimal target heading ${\theta}_{i}^{g_*}$ (Lines 10-12).

\subsection{Online trajectory prediction}
In this paper, we use the energy function \eqref{E26_headenergycost} and target headings ${\theta}^{g_*} $ to forecast agents' future trajectories. Based on agents' current states at $t$, we calculate their approximate optimal velocity by \textbf{Algorithm} \ref{Alg2_theta_opt} and update their positions. With ${\rm N}_p$ iterations in the time horizon, we acquire the agents' predicted future trajectories.

\begin{algorithm}[htb]
\label{Alg5_OnlineTrajPred}
\caption{\texttt{Trajectory\_prediction}}
\KwIn{Observed positions $\{{\bf{P}}_{t}^i\}$, number of agents ${\rm{M}}_t$, static environment states ${\mathbf P}_{obst}$}
\KwOut{Multi-agent predicted trajectories $\{ {{\bf{\hat P}}^i}\} $}

$\{G_l\}$, $\{u_l\}$=\texttt{Group\_information}($\{{\bf{P}}_{ob}^i\}$, ${\rm{M}}_t$)\;

\For{$i=[1,2,...,{\rm{M}}_t]$}
{
$\Theta_i^*$ = \texttt{Opt\_parameters\_estimation}(\{\{${\bf{P}}_{ob}^i$\}, ${{\mathbf P}_{obst}}$\}, $\{G_l\}$, $\{u_l\}$)\;
${\theta}_{i}^{g_*}$ = \texttt{Target\_heading\_estimation}($\Theta_i^*$, \{\{${\bf{P}}_{ob}^i$\}, ${{\mathbf P}_{obst}}$\})\;
}
Initial the agent's current state ${\bf{\hat p}}_t^i = {\bf p}_t^i$ and velocity ${\bf{\hat v}}_t^i = ({\bf{p}}_t^i - {\bf{p}}_{t - 1}^i)/\Delta t$ with the observed state ${\bf{P}}_i$\;

\For{$k=[1,2,...,{\rm{N}}_p]$}
{
\For{$i=[1,2,...,{\rm{M}}_t]$}
{
Calculate the values ${\bf A}^i$ in \eqref{E26_headenergycost} with estimated information\;

${{\bf{\hat v}}_{t+k}^i}$ = \texttt{Opt\_velocity\_estimation}(${\bf{\hat v}}_{t+k-1}^i$, $\Theta_i^*$, ${\bf A}^i$)\;

Update ${\bf{\hat p}}_{t+k}^i$ by \eqref{F13_posupdate} with the velocity ${{\bf{\hat v}}_{t+k}^i}$ and the state ${\bf{\hat p}}_{t+k-1}^i$\;
Add the ${\bf{\hat p}}_{t+k}^i$ to ${{\bf{\hat P}}^i}$\;
}
}
\end{algorithm}
The details of online trajectory prediction are shown in \textbf{Algorithm} \ref{Alg5_OnlineTrajPred}. Firstly, Line 1 divides agents' groups by \textbf{Algorithm} \ref{Alg1_group} with observed agents' positions $\{{\bf{P}}_t^i\}$. Secondly, we estimate approximate optimal parameters and target headings by \textbf{Algorithm} \ref{Alg2_theta_opt} and \textbf{Algorithm} \ref{Alg4_Target_Head} (Lines 2-4). Then, based on the estimated information and agents' current states (Line 5), we calculate the pre-calculated value ${\bf A}^i$ at the predicted time $t+k$, $k \ge 1$ (Line 8). Next, the future velocity of agent $i$ is predicted by \textbf{Algorithm} \ref{Alg3_V_opt} (Line 9). Note that the evaluation function \eqref{E12_newEnergy} in Line 4 of \textbf{Algorithm} \ref{Alg3_V_opt} is replaced by \eqref{E26_headenergycost} in this process. Finally, we update agents' future states to get predicted trajectories (Lines 10-11).

\section{Experiment Results}\label{Sec5_results}
This section firstly presents the way of evaluating the accuracy of predicted trajectories. Then it introduces datasets in the experiments. Finally, we conduct experiment and comparison studies to verify the effectiveness and advantages of the proposed trajectory prediction method.

\subsection{Predicted trajectory evaluation}
There are several methods to evaluate the errors between the predicted trajectory and the ground truth. The two most popular evaluation metrics are average displacement error (ADE) and final displacement error (FDE) \cite{Golnaz2021, Liang2020,Pang2021}. They are defined as:
\begin{subequations}
\begin{gather}
{\rm{ADE = }}\frac{1}{{{{\rm{N}}_{eva}} \times {\rm{N}}_p}}\sum\limits_{i = 1}^{{\rm{N}}_{eva}} {\sum\limits_{k = 1}^{{{\rm{N}}_p}} {d\left( {{\bf{\hat p}}_{t+k}^i,{\bf{p}}_{t+k}^i} \right)} }  ,\label{E30a_ADE}\\
{\rm{FDE = }}\frac{1}{{\rm{N}}_{eva}}\sum\limits_{i = 1}^{{{\rm{N}}_{eva}}} {d\left( {{\bf{\hat p}}_{t+{{\rm{N}}_p}}^i,{\bf{p}}_{t+{{\rm{N}}_p}}^i} \right)} ,\label{E30b_FDE}
\end{gather}
\end{subequations}
where ${\rm N}_{eva}$ is the total number of agents in the prediction process.

\begin{figure*}
	\centering
	\includegraphics[width=14cm]{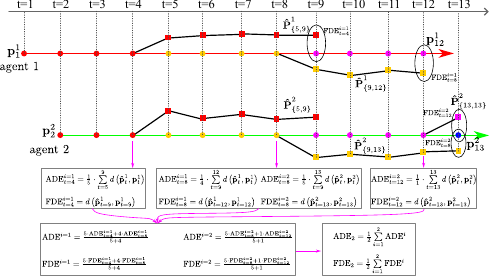}	 
	\caption{Predicted trajectory evaluation. The red and cyan lines with dots are agents' real trajectories, the dots with the same colors represent the observed states in one prediction process, the black lines with squares are predicted trajectories, ${\bf p}_1^1$ and ${\bf p}_2^2$ are agents' start positions, and ${\bf p}_{12}^1$ and ${\bf p}_{13}^2$ are their end positions. The future trajectories are predicted at $t=4$, $t=8$, and $t=12$ respectively. The observation length is ${\rm N}=4$, the prediction length is ${\rm N}_p = 5$, and the threshold $\Delta{\rm N}_{ob} = 4$.}
	\label{F5_predTrajEvalu}
\end{figure*}

However, in some studies such as \cite{Golnaz2021, Liang2020,Pang2021}, they sampled 20 predicted trajectories by the prediction methods and selected the one that has the minimum value in \eqref{E30a_ADE} and \eqref{E30b_FDE} from the samples. In addition, the predicted trajectory of agent $i$ is just evaluated one time, even though the trajectory length of agent $i$ is larger than ${\rm{N}}_p$. This prevents the predicted trajectories from being used in some tasks that need accurate real-time predicted trajectories, such as robot path planning and target tracking. In their results, one cannot know which predicted future trajectory is the most accurate one at the current time. In contrast, our method deals with the agents' trajectories during the whole active states of agents and predicts one trajectory for agent $i$ in each observation period. We calculate ADE and FDE by
\begin{subequations}
\begin{gather}
{\rm{AD}}{{\rm{E}}_2}{\rm{ = }}\frac{1}{{{{\rm{N}}_{eva}}}}\sum\limits_{i = 1}^{{{\rm{N}}_{eva}}} {\left( {\frac{{\sum\limits_{l = 1}^{{{\rm{K}}_i}} {\sum\limits_{k = 1}^{{\rm{N}}_l^i} {d\left( {{\bf{\hat p}}_{t + k}^i,{\bf{p}}_{t + k}^i} \right)} } }}{{{\rm{N}}_1^i + {\rm{N}}_2^i +  \cdots  + {\rm{N}}_{{{\rm{K}}_i}}^i}}} \right)}  ,\label{E31_ADE2}\\
{\rm{FD}}{{\rm{E}}_2}{\rm{ = }}\frac{1}{{{\rm{N}}_{eva}}}\sum\limits_{i = 1}^{{\rm{N}}_{eva}} {\left( {\frac{{\sum\limits_{l = 1}^{{{\rm{K}}_i}} {\left( {{\rm{N}}_l^i \cdot d\left( {{\bf{\hat p}}_{t+{{\rm{N}}_l^i}}^i,{\bf{p}}_{{t+{\rm N}_l^i}}^i} \right)} \right)} }}{{{\rm N}_1^i + {\rm{N}}_2^i +  \cdots  + {\rm{N}}_{{{\rm{K}}_i}}^i}}} \right)} ,\label{E32_FDE2}
\end{gather}
\end{subequations}
where ${\rm{K}}_i$ is the number of prediction times in which corresponding lengths of observation are not less than $\Delta{\rm N}_{ob}$, and ${\rm{N}}_l^i  \le {\rm N}_p$ is the compared length between the predicted trajectory and the real trajectory. Different from \eqref{E30a_ADE} and \eqref{E30b_FDE}, we do not evaluate the predicted trajectories whose observation length ${\rm N} < \Delta{\rm N}_{ob}$, because the predicted trajectories with a small number of observed states have large errors. \mbox{Fig. \ref{F5_predTrajEvalu}} shows an example of the predicted trajectory evaluation process. Firstly, at time $t=4$, the observed states are $\left\{ {{\bf{p}}_1^1,{\bf{p}}_2^1,{\bf{p}}_3^1,{\bf{p}}_4^1} \right\}$ and $\left\{ {{\bf{p}}_2^2,{\bf{p}}_3^2,{\bf{p}}_4^2} \right\}$, the predicted trajectories are ${\bf{\hat P}}_{\{ 5,9\} }^1$ and ${\bf{\hat P}}_{\{ 5,9\} }^2$. When evaluating the predicted trajectories at $t=4$, there are three states in observed states of agent 2, which is $3 <\Delta{\rm N}_{ob} = 4$. In this case, we do not consider the predicted errors of agent 2, and the average displacement error and final displacement error of agent 1 are ${\rm{ADE}}_{t = 4}^{i = 1}$ and ${\rm{FDE}}_{t = 4}^{i = 1}$. Then, from $t=5$ to $t=8$, the newly observed states are shown in the figure with yellow dots, and ${\bf{\hat P}}_{\{ 9,12\} }^1$ and ${\bf{\hat P}}_{\{ 9,13\} }^2$. The comparison length of the agent 1 is ${\rm N}_{l=2}^{i=1} = 4$, because the agent 1 gets out of the visual field at $t=12$. Finally, there is only one predicted trajectory ${\bf{\hat P}}_{\{ 13,13\} }^2$, and the predicted errors are ${\rm{ADE}}_{t = 12}^{i = 2}$ and ${\rm{FDE}}_{t = 12}^{i = 2}$. Consequently, the ${\rm ADE}_2$ and ${\rm FDE}_2$ are calculated based on ${\rm ADE}_t^i$ and ${\rm FDE}_t^i$, and the detail calculation is shown in \mbox{Fig. \ref{F5_predTrajEvalu}}.

\subsection{Datasets of test}
To evaluate the proposed method, we use pedestrian walking datasets of ETH, HOTEL, UNIV, ZARA1 and ZARA2 in \cite{Pellegrini2009,Lerner2007} to test the proposed algorithm. \mbox{Table \ref{Tab1_datasetInf}} summarizes the basic information of datasets. {In ETH, the scene is looked down by the camera from the building, and pedestrians walk across a thoroughfare where its sides are grasslands. Moreover, the area where pedestrians in ETH have appeared is the widest. The data of HOTEL recorded pedestrians walking on the road that has a bus station in the front, and the pedestrians may stay there for the bus; the changing of the environment structure will influence prediction accuracy. The average density and maximum density (pedestrians/frame) represent the crowded degree of pedestrians' walking space; the environment of UNIV is the most crowded and complex, which has more interaction with pedestrians. ZARA1 and ZARA2 are in the same environment, and the pedestrians' motions are also influenced by the vehicles.

\begin{table}[htbp]
  \centering
  \caption{The information of datasets. The unit of $\Delta {\rm X}$ and $\Delta {\rm Y}$, average density and maximum density, average speed are meter, pedestrian/frame, meter/second respectively.}
  \renewcommand\arraystretch{1.1}
    \begin{tabular}{ccccccc}
    \toprule
    \multicolumn{2}{c}{Dataset} & ETH   & HOTEL & UNIV  & ZARA1 & ZARA2 \\
    \midrule
    \multirow{2}[3]{*}{Area} & $\Delta {\rm X}$  & 16.6  & 14.6  & 11.5  & 11.9  & 13 \\
\cmidrule{2-2}          & $\Delta {\rm Y}$ & 21.3  & 7.7   & 16.6  & 18.5  & 18.9 \\
    \multicolumn{2}{c}{Pedestrians} & 360   & 390   & 434   & 148   & 204 \\
    \multicolumn{2}{c}{Frames} & 1448  & 1168  & 541   & 866   & 1052 \\
    \multicolumn{2}{c}{Average density} & 6.2   & 5.6   & 32.5  & 5.8   & 9.1 \\
    \multicolumn{2}{c}{Maximum density} & 27    & 18    & 51    & 20    & 18 \\
    \multicolumn{2}{c}{Average speed} & 1.43  & 1.15  & 0.8   & 1.4   & 1.36 \\
    \bottomrule
    \end{tabular}%
  \label{Tab1_datasetInf}%
\end{table}%

In these datasets, the interval time $\Delta t$ between adjacent frames is $0.4$s. We first sample up to ${\rm N}=8$ states of pedestrians and then predict their ${\rm N}_p = 12$ future states.

\subsection{Experiment analysis and results}
To evaluate the effectiveness of the proposed group division method, we set the $\Delta {\delta _{{\rm{dF}}}} = 1.8$ in \eqref{E19_similaritymatrix}. From \mbox{Fig. \ref{F6_Groupets_eth}} and \mbox{Fig. \ref{F7_Groupets_hotel}}, it can be seen that the groups of pedestrians in datasets can be divided correctly at most times. The average accuracy of the group division in ETH and HOTEL is $0.815$ and $0.879$, respectively. Compared with the group prediction precision in \cite{Yamaguchi2011}, the corresponding values are $0.820$ and $0.885$ with the observed length ${\rm N}=8$ which is similar to our results. Note that we do not always observe pedestrians' states with the length ${\rm N}=8$ as mentioned above. In this case, using the different lengths of states to divide the group will affect the accuracy of the result. In the group division by the Frechet distance method, there is only one freedom parameter $\Delta {\delta _{{\rm{dF}}}}$, which needs to be considered. It is more efficient compared with the method in \cite{Yamaguchi2011}, which needs to consider the distances, velocities, moving headings {\em etc}., and trained by SVM. \mbox{Fig. \ref{F8_grouptraject}} shows the trajectories in the same groups estimated by the Frechet distance method. It is obvious that the motions of pedestrians in the estimated groups are influenced by the sub-energy function ${\rm E}_{att}(\cdot)$ in \eqref{E5_attractfunction}.
\begin{figure}[htb]
	\centering
	\subfigure[]{
		\includegraphics[width=8.4cm]{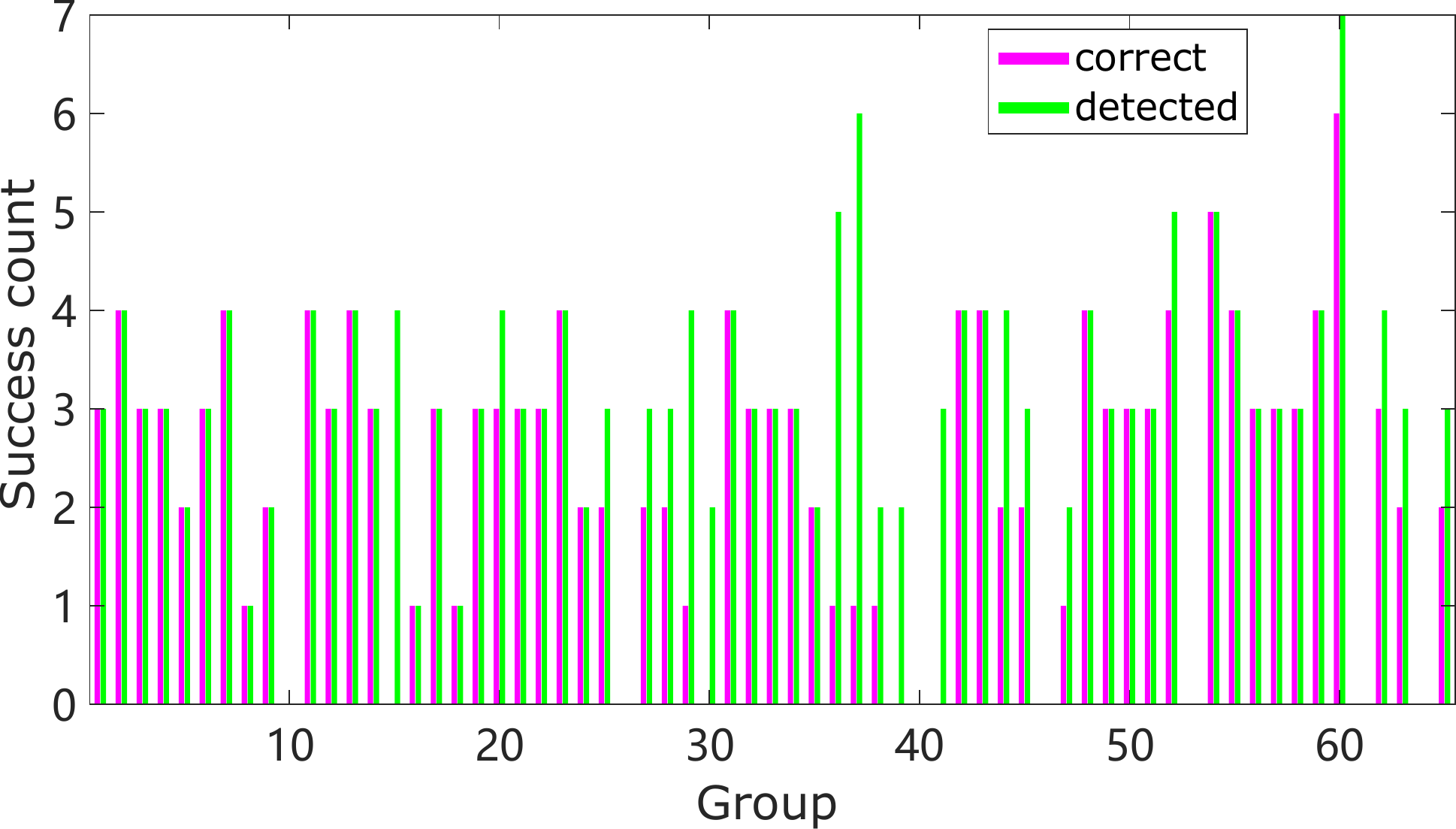} 
		\label{F6_Groupets_eth}
	}
	\quad
	\subfigure[]{
		\includegraphics[width=8.4cm]{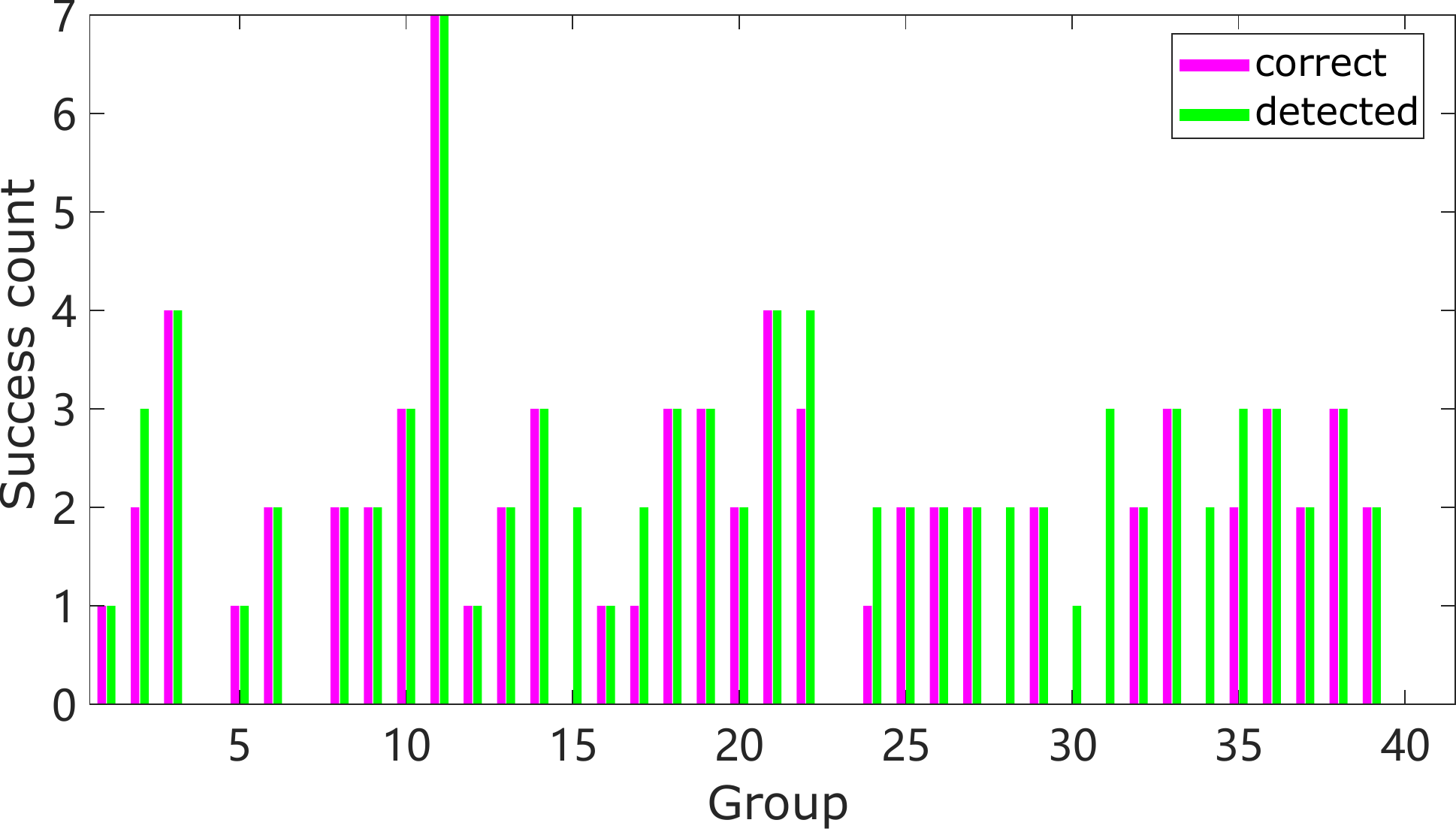}	
		\label{F7_Groupets_hotel}
	}
	\caption{Number of wholes detected groups and correct groups, the cyan bar represents the observed times of elements in the real group, and the red bar represents the correct division times in the detected groups. (a) The group division of ETH, the total number of real groups is 65. (b) The group division of HOTEL, the total number of real groups is 41.}
\end{figure}

\begin{figure}[htb]
	\centering
   \def\svgwidth{8.4cm}
	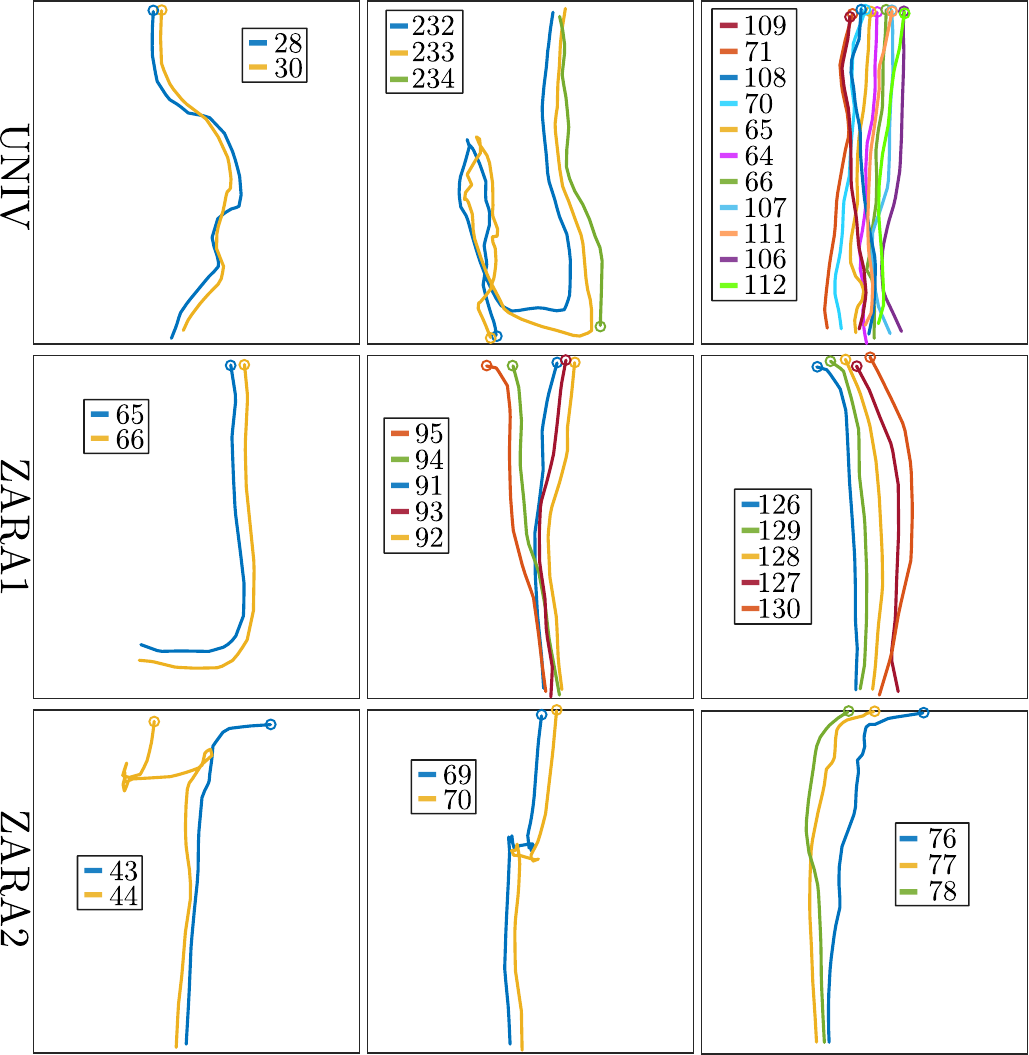
	\caption{The trajectories of estimated groups. The circles are the start positions of trajectories.}
	\label{F8_grouptraject}
\end{figure}

\begin{table*}[htbp]
  \centering
  \caption{Average displacement error and final displacement error of predicted trajectories, and the unit of ADE and FDE are the meter. In evaluation, we set $\Delta{\rm N}_{ob} = 7$ in \eqref{E31_ADE2} and \eqref{E32_FDE2} which observing 8 frames and predicting next 12 frames.}

    \begin{tabular}{ccccccccc}
    \toprule
          & Method & ETH   & HOTEL & UNIV  & ZARA1 & ZARA2 & AVG   & Evaluation method \\
    \midrule
    \multirow{3}[2]{*}{Learning-based} & SimFuse \cite{Golnaz2021} & 0.59/1.18 & 0.31/0.73 & 0.50/1.17 & 0.27/0.54 & 0.27/0.56 & 0.38/0.84 & \multirow{3}[2]{*}{the best one in 20 samples} \\
          & LSTM-Based GAN \cite{Sun2022} & 0.69/1.24 & 0.43/0.87 & 0.53/1.17 & 0.28/0.61 & 0.28/0.59 & 0.44/0.90 &  \\
          & Social GAN \cite{Agrim2018} & 0.87/1.62 & 0.67/1.37 & 0.76/1.52 & 0.35/0.68 & 0.42/0.84 & 0.61/1.21 &  \\
\cmidrule{1-1}\cmidrule{9-9}    \multicolumn{1}{l}{Physics-based} & Linear & 1.33/2.94 & 0.39/0.72 & 0.82/1.59 & 0.62/1.21 & 0.77/1.48 & 0.79/1.59 & \multirow{4}[4]{*}{one sample} \\
\cmidrule{1-1}    \multirow{3}[2]{*}{Planning-based} & RVO   & 0.63/0.92 & 0.65/0.58 & 1.24/2.11 & 0.65/1.34 & 1.11/1.08 & 0.86/1.21 &  \\
          & The proposed approach & \multirow{2}[1]{*}{0.45/0.90} & \multirow{2}[1]{*}{0.32/0.60} & \multirow{2}[1]{*}{0.62/1.32} & \multirow{2}[1]{*}{0.46/1.01} & \multirow{2}[1]{*}{0.57/1.21} & \multirow{2}[1]{*}{0.48/1.01} &  \\
          & in this work &       &       &       &       &       &       &  \\
    \bottomrule
    \end{tabular}%
  \label{T2_compared}%
\end{table*}%

\begin{figure}[htb]
	\centering
	\includegraphics[width=8.6cm]{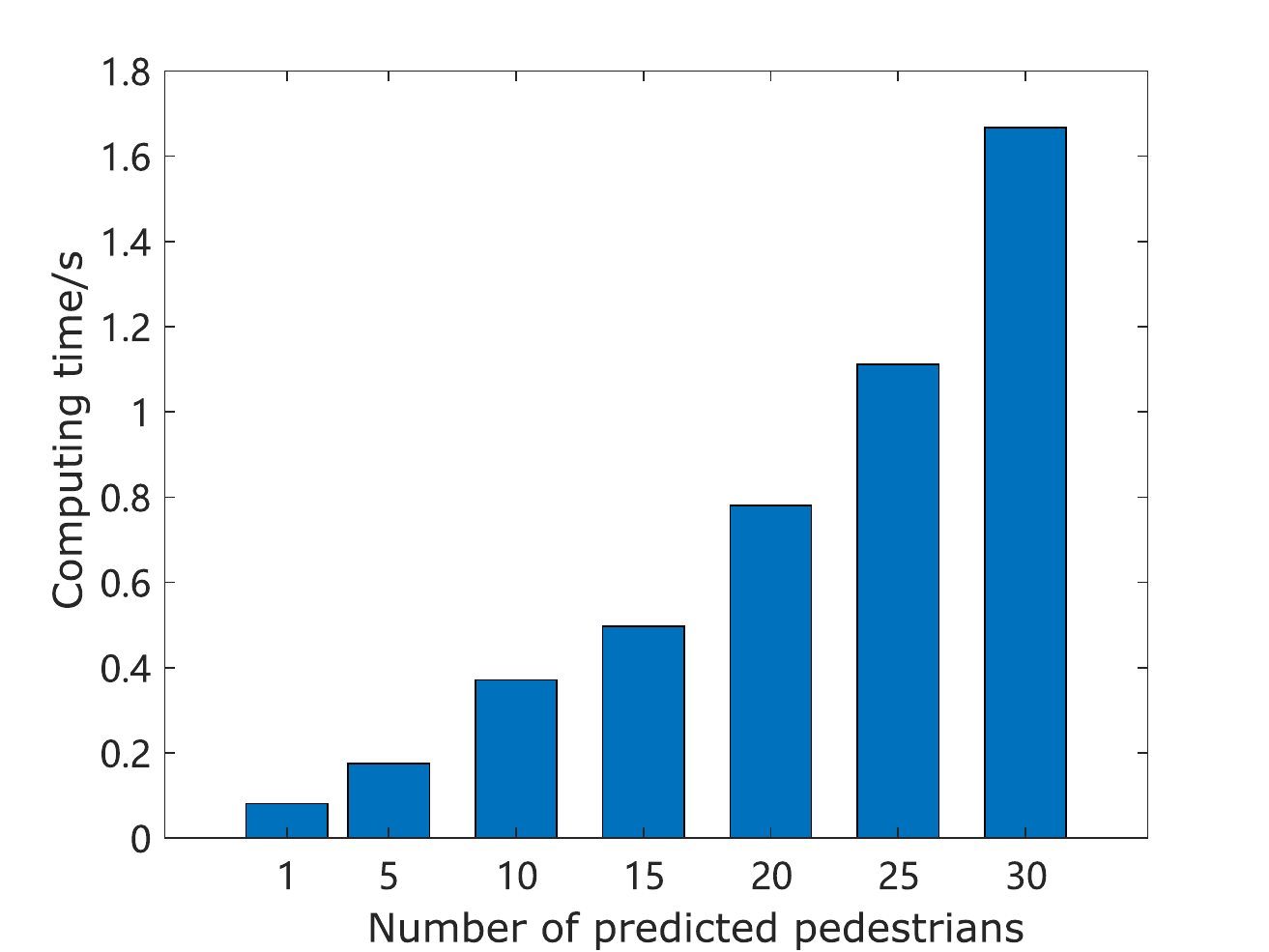}	
	\caption{The computing time with the number of predicted pedestrians. There are ${\rm N}_{\Theta}=12$ sampled salps and ${\rm L}=10$ iterations in \textbf{Algorithm} \ref{Alg2_theta_opt}. In \textbf{Algorithm} \ref{Alg3_V_opt}, the numbers of sample salps and iterations are ${\rm N_v}=10$ and ${\rm L_v}=5$, respectively. The number of sampled target headings is ${\rm N}_{\theta} = 31$ in \textbf{Algorithm} \ref{Alg4_Target_Head}. The computation platform is Intel(R) Core(TM) i7-10700F CPU $@$ 2.90GHz by Qt Creator with the C language.}
	\label{Fig9_1_consumeTime}
\end{figure}

To verify the advantages of the proposed method, we investigate the state-of-the-art trajectory prediction methods in \cite{Golnaz2021,Sun2022,Agrim2018} and the RVO method to compare with our results. From \mbox{Table \ref{T2_compared}}, the errors of linear and RVO methods are the worst. The linear method does not fuse agents' interactions and the environment; it merely considers the self motions. Although the RVO method considers agents' interactions, the constant interaction characteristics make it unable to adapt to complex dynamic environments such as the UNIV. Our method has minimum errors in ETH and the minimum FDE in HOTEL. The methods of SimFuse, LSTM-based GAN, and Social GAN have better performances in UNIV, ZARA1, and ZARA2, but it depends on selecting the minimum evaluation error from multiple samples. For the proposed method, only one predicted trajectory is used to evaluate its errors of ADE and FDE. \mbox{Fig. \ref{Fig9_1_consumeTime}} displays the computing time of the proposed prediction method. Within 0.8s, the method can predict 20 pedestrians' future trajectories. The predicted errors and computing time show the advantages of the proposed method in terms of the accuracy and real-time performance.

\begin{figure*}
	\centering
	\includegraphics[width=15cm]{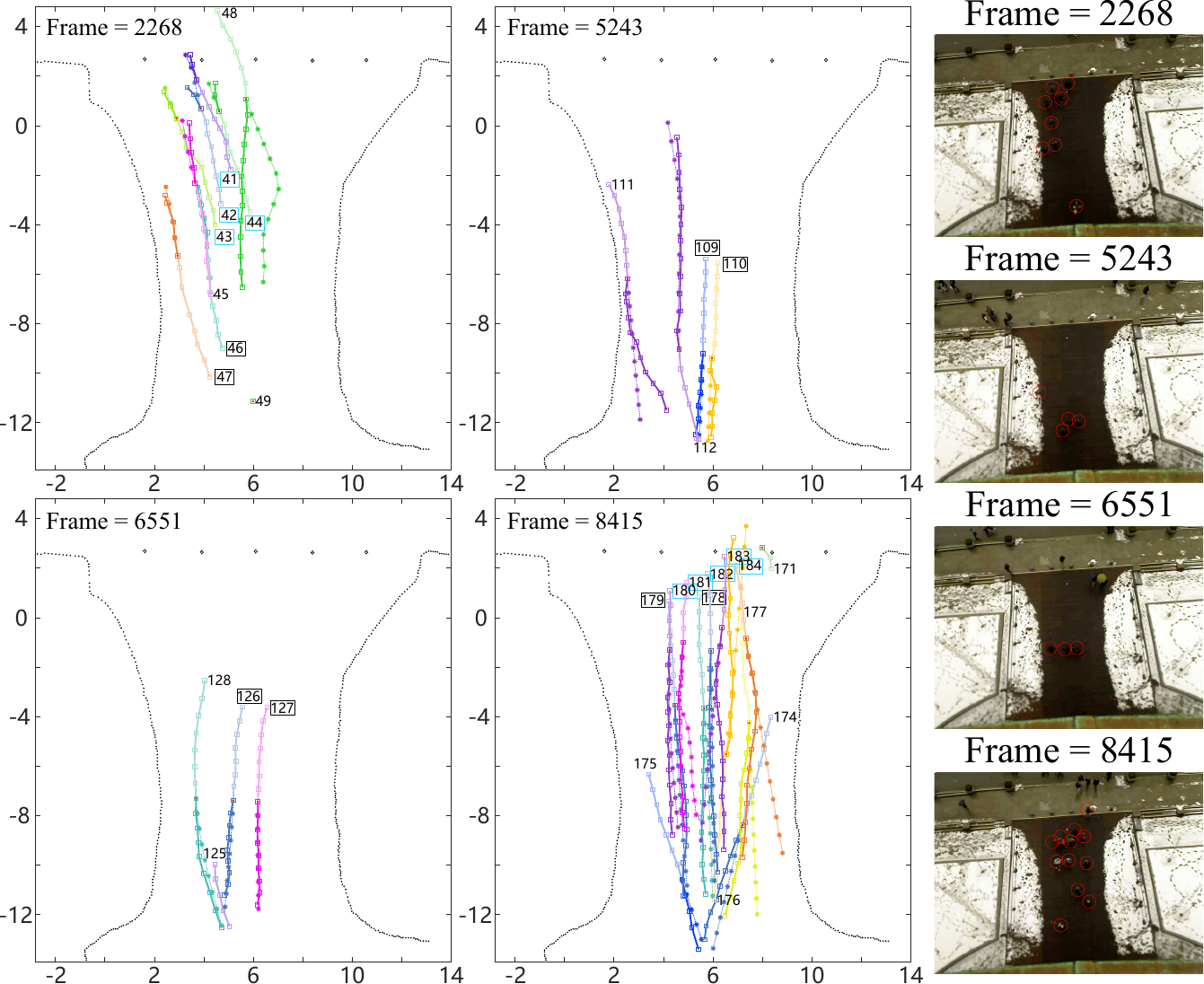}	 	
	\caption{The predicted trajectories in ETH. The squares with light and dark colors are observed states and real future trajectory nodes of pedestrians, respectively; the dotted lines with the marks of * are predicted trajectories; the red dots in squares represent pedestrians' current positions; the black points are discrete environment obstacles. The pedestrian IDs in the same color boxes indicate they are in one group.}
	\label{Fig9_ETH_predT}
\end{figure*}

To further illustrate the influences of factors, such as group attraction, interaction sensitivity, crowded environment and target heading, the predicted trajectories of datasets are shown in \mbox{Fig. \ref{Fig9_ETH_predT}}, \mbox{Fig. \ref{Fig10_HOTEL_pred}}, \mbox{Fig. \ref{Fig11_UNIV_predT}} and \mbox{Fig. \ref{Fig12a_zarapred}}.

\mbox{Fig. \ref{Fig9_ETH_predT}} displays the real and predicted trajectories in the environment of ETH. The group attractions improve the prediction accuracy from the frames of 2268, 5243, 6551 and 8415. When considering the influence of environment's static obstacles, the predicted trajectory of pedestrian 48 in frame 2268 has a small FDE. By using the proposed target headings estimation method in \textbf{Algorithm} \ref{Alg4_Target_Head}, the predicted states of pedestrians 112 and 128 are highly similar to their real trajectories. In frame 8415, there are complex interactions in the crowd, however, the predicted trajectories still have small errors with 0.37 and 0.71 of ADE and FDE, respectively.

\begin{figure*}[htb]
	\centering
	\subfigure[]{
		\includegraphics[width=8.5cm]{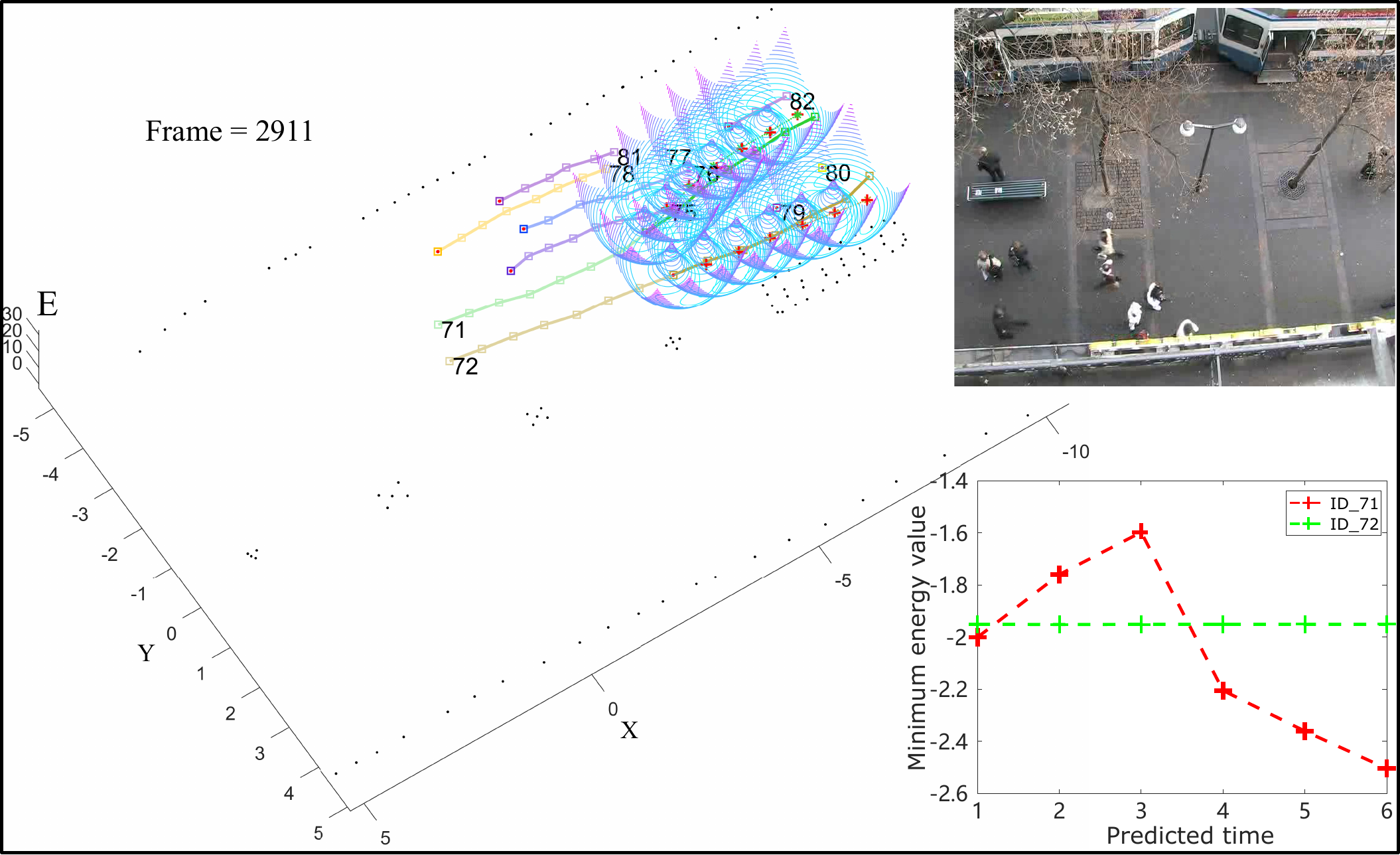}
		\label{Fig10a_HOTEL_pred}	
	}
	\quad
	\subfigure[]{
		\includegraphics[width=8.5cm]{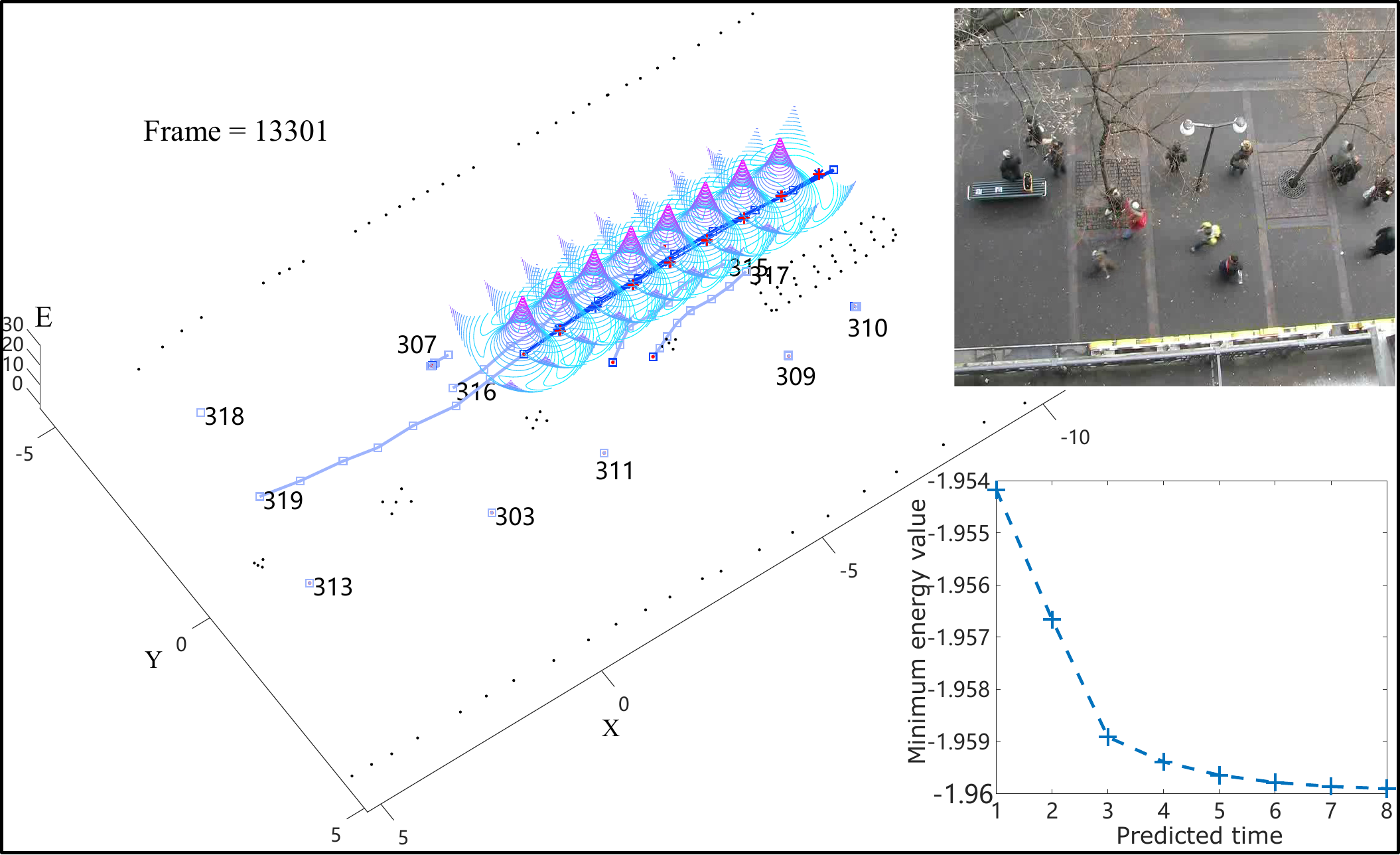}	
		\label{Fig10b_HOTEL_pred}
	}
	\caption{Predicted trajectories and their energy cost distributions of HOTEL. The XY plane represents the moving space of pedestrians; E axis is the value of energy costs, 
which using the function ${{\rm{E}}_{\Theta _i^*}}(\cdot)$ to calculate the costs of selected velocities, then mapping the velocities and their corresponding values to XYE space with motion interval $\Delta t = 0.4s$; the red marks $+$ are positions which have minimum energy cost values; the line charts in the left lower corner are the changing of predicted nodes' minimum energy values.}
	\label{Fig10_HOTEL_pred}
\end{figure*}

\mbox{Fig. \ref{Fig10_HOTEL_pred}} displays energy values' distributions of the predicted trajectories' nodes. In \mbox{Fig. \ref{Fig10a_HOTEL_pred}}, pedestrians 71 and 72 belong to one group, and their optimal sets are $\Theta _{71}^* = \{ 0.14,6.86,1.96,0.49,0.02,\fbox{0.18,4.81,2.14}\} $ and $\Theta _{72}^* = \{ 0.14,6.86,1.96,0.01,0.78,\fbox{0.02,0.10,0.00}\} $. Compared with the values of ${{\rm{w}}_i}$, ${d_i}$, ${\alpha _i}$ (the values in black boxes) in $\Theta _{71}^*$ and $\Theta _{72}^*$, it is obvious that pedestrian 71 is more sensitive to other pedestrians and the environment's obstacles. The real trajectories show the significance of estimated $\Theta _{71}^*$ and $\Theta _{72}^*$ at the same time. The static pedestrians 78 and 80 are in the middle of pedestrians 71 and 72; pedestrian 72 is barely affected by 78 and 80, and pedestrian 71 moves far from them in contrast. The estimated target heading ${\theta}_{319}^{g_*}$ of pedestrian 319 in \mbox{Fig. \ref{Fig10b_HOTEL_pred}} is $-178.71^{\circ}$, and the optimal set is $\Theta _{319}^* = \{ 0.14,6.86,1.96,0.00,0.00,0.98,0.10,0.00\}$; the values mean that the motion of pedestrian 319 is mainly controlled by its self behaviors.

\begin{figure*}[htb]
	\centering
	\includegraphics[width=16cm]{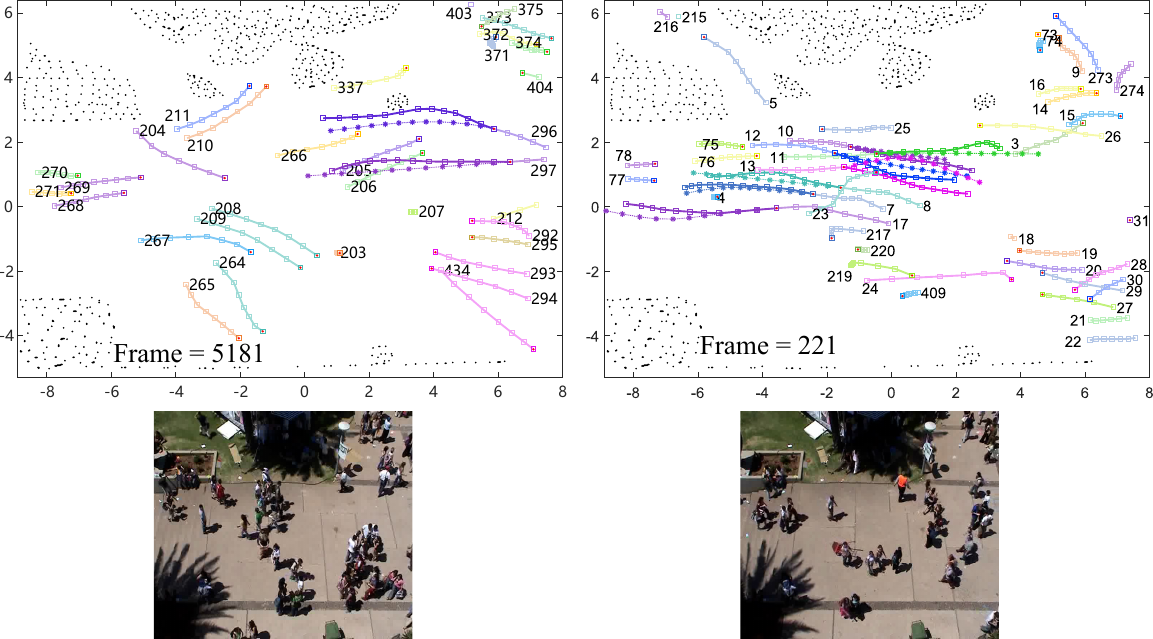}	
	\caption{Predicted trajectories of UNIV. In frame 221, pedestrians 10, 11, 12 and 13 are in one group, pedestrians 7, 8 and 17 are in another group; in frame 5181, pedestrians 296 and 291 belong to the same group.}
	\label{Fig11_UNIV_predT}
\end{figure*}

The scene of the UNIV is the most complex and crowded in the five datasets, which can be seen from \mbox{Fig. \ref{Fig11_UNIV_predT}}. The complex coupled interactions of pedestrians account for the low accuracy of predicted trajectories; however, our method still performs well than methods of linear, Social GAN and RVO. 

\begin{figure*}[htb]
	\centering
	\subfigure[]{
		\includegraphics[height=6.5cm]{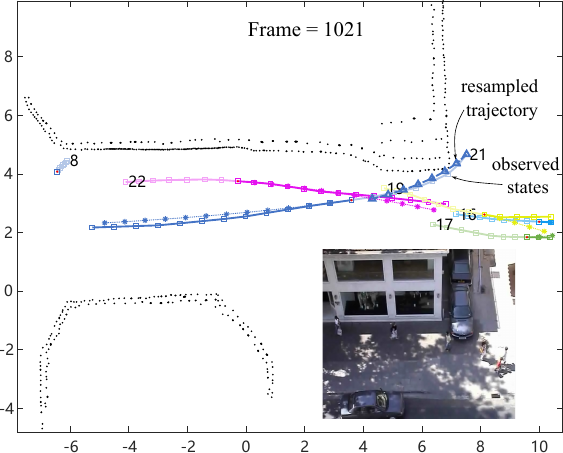}
		\label{Fig12a_zarapred}	
	}
	\quad
	\subfigure[]{
		\includegraphics[height=6.5cm]{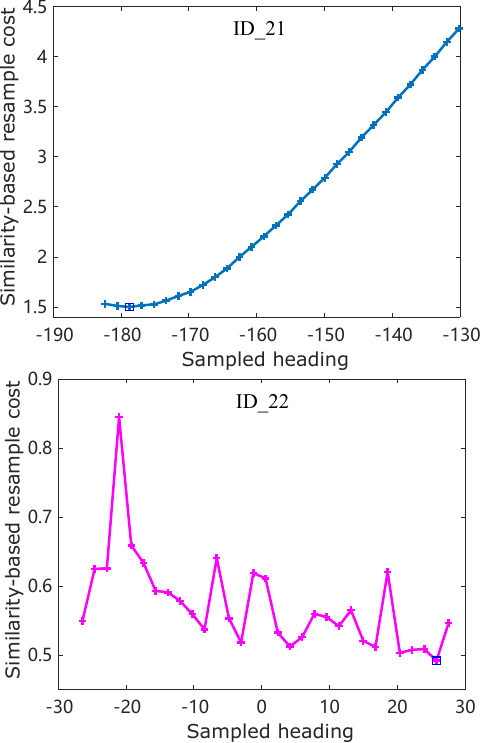}	
		\label{Fig12b_zarapred}
	}
	\caption{Predicted trajectories and sampled target headings' cost values of ZARA1. (a) Predicted trajectories, the dotted line with triangles is the resampled trajectory by the estimated optimal target heading of pedestrian 21. (b) Sampled target headings and their similarity-based resample costs, the blue squares are the estimated optimal target headings of pedestrians 21 and 22, respectively.}
\end{figure*}

The datasets of ZARA1 and ZARA2 are in the same environment. In \mbox{Fig. \ref{Fig12a_zarapred}}, the target heading of pedestrian 21 is $-178.73^{\circ}$, and its corresponding resampled trajectory is similar and close to the real observed trajectory. As for the estimated target heading of pedestrian 22, although it deviates from the true value, the errors of the predicted trajectory are still small in comparison with the value of the energy function.

\section{Conclusion}\label{Sec6_conclusion}
In this work, we have presented the high-precision trajectory prediction method for multiple agents with the energy function model in crowded scenes. The Frechet distance has been proposed to divide groups of observed agents. The efficient decoupled cost function and SSA-GD algorithms are designed to solve the optimization problem. The SRE method has been proposed to estimate target headings of agents. Experiment and comparison based on standard datasets have been conducted to confirm the advantages and performance of the proposed method. In the future studies, we will analyze the error caused by simplifying the optimization function \eqref{E15_trainpara} with \eqref{E23_newOpt}, explore the theoretical analysis of the algorithm performance and solve the trajectory prediction with multiple sensors in partially observed environments.

\input{Trajectory_Prediction_Similarity_Based_TXY_RsN1.bbl}

\end{document}

%% file: interaction_multi.pdf_tex
\begingroup%
  \makeatletter%
  \providecommand\color[2][]{%
    \errmessage{(Inkscape) Color is used for the text in Inkscape, but the package 'color.sty' is not loaded}%
    \renewcommand\color[2][]{}%
  }%
  \providecommand\transparent[1]{%
    \errmessage{(Inkscape) Transparency is used (non-zero) for the text in Inkscape, but the package 'transparent.sty' is not loaded}%
    \renewcommand\transparent[1]{}%
  }%
  \providecommand\rotatebox[2]{#2}%
  \newcommand*\fsize{\dimexpr\f@size pt\relax}%
  \newcommand*\lineheight[1]{\fontsize{\fsize}{#1\fsize}\selectfont}%
  \ifx\svgwidth\undefined%
    \setlength{\unitlength}{412.06109475bp}%
    \ifx\svgscale\undefined%
      \relax%
    \else%
      \setlength{\unitlength}{\unitlength * \real{\svgscale}}%
    \fi%
  \else%
    \setlength{\unitlength}{\svgwidth}%
  \fi%
  \global\let\svgwidth\undefined%
  \global\let\svgscale\undefined%
  \makeatother%
  \begin{picture}(1,0.56270378)%
    \lineheight{1}%
    \setlength\tabcolsep{0pt}%
    \put(0,0){\includegraphics[width=\unitlength,page=1]{interaction_multi.pdf}}%
  \end{picture}%
\endgroup%

%% file: obst_headselect.pdf_tex
\begingroup%
  \makeatletter%
  \providecommand\color[2][]{%
    \errmessage{(Inkscape) Color is used for the text in Inkscape, but the package 'color.sty' is not loaded}%
    \renewcommand\color[2][]{}%
  }%
  \providecommand\transparent[1]{%
    \errmessage{(Inkscape) Transparency is used (non-zero) for the text in Inkscape, but the package 'transparent.sty' is not loaded}%
    \renewcommand\transparent[1]{}%
  }%
  \providecommand\rotatebox[2]{#2}%
  \newcommand*\fsize{\dimexpr\f@size pt\relax}%
  \newcommand*\lineheight[1]{\fontsize{\fsize}{#1\fsize}\selectfont}%
  \ifx\svgwidth\undefined%
    \setlength{\unitlength}{453.54330709bp}%
    \ifx\svgscale\undefined%
      \relax%
    \else%
      \setlength{\unitlength}{\unitlength * \real{\svgscale}}%
    \fi%
  \else%
    \setlength{\unitlength}{\svgwidth}%
  \fi%
  \global\let\svgwidth\undefined%
  \global\let\svgscale\undefined%
  \makeatother%
  \begin{picture}(1,0.4625)%
    \lineheight{1}%
    \setlength\tabcolsep{0pt}%
    \put(0,0){\includegraphics[width=\unitlength,page=1]{obst_headselect.pdf}}%
  \end{picture}%
\endgroup%

%% file: groupTrajectory.pdf_tex
\begingroup%
  \makeatletter%
  \providecommand\color[2][]{%
    \errmessage{(Inkscape) Color is used for the text in Inkscape, but the package 'color.sty' is not loaded}%
    \renewcommand\color[2][]{}%
  }%
  \providecommand\transparent[1]{%
    \errmessage{(Inkscape) Transparency is used (non-zero) for the text in Inkscape, but the package 'transparent.sty' is not loaded}%
    \renewcommand\transparent[1]{}%
  }%
  \providecommand\rotatebox[2]{#2}%
  \newcommand*\fsize{\dimexpr\f@size pt\relax}%
  \newcommand*\lineheight[1]{\fontsize{\fsize}{#1\fsize}\selectfont}%
  \ifx\svgwidth\undefined%
    \setlength{\unitlength}{493.61330047bp}%
    \ifx\svgscale\undefined%
      \relax%
    \else%
      \setlength{\unitlength}{\unitlength * \real{\svgscale}}%
    \fi%
  \else%
    \setlength{\unitlength}{\svgwidth}%
  \fi%
  \global\let\svgwidth\undefined%
  \global\let\svgscale\undefined%
  \makeatother%
  \begin{picture}(1,1.02570065)%
    \lineheight{1}%
    \setlength\tabcolsep{0pt}%
    \put(0,0){\includegraphics[width=\unitlength,page=1]{groupTrajectory.pdf}}%
  \end{picture}%
\endgroup%

%% file: Trajectory_Prediction_Similarity_Based_TXY_RsN1.bbl

%% file: Trajectory_Prediction_Similarity_Based_TXY_RsN1.bbl
\begin{thebibliography}{10}
\providecommand{\url}[1]{#1}
\csname url@samestyle\endcsname
\providecommand{\newblock}{\relax}
\providecommand{\bibinfo}[2]{#2}
\providecommand{\BIBentrySTDinterwordspacing}{\spaceskip=0pt\relax}
\providecommand{\BIBentryALTinterwordstretchfactor}{4}
\providecommand{\BIBentryALTinterwordspacing}{\spaceskip=\fontdimen2\font plus
\BIBentryALTinterwordstretchfactor\fontdimen3\font minus
  \fontdimen4\font\relax}
\providecommand{\BIBforeignlanguage}[2]{{%
\expandafter\ifx\csname l@#1\endcsname\relax
\typeout{** WARNING: IEEEtranS.bst: No hyphenation pattern has been}%
\typeout{** loaded for the language `#1'. Using the pattern for}%
\typeout{** the default language instead.}%
\else
\language=\csname l@#1\endcsname
\fi
#2}}
\providecommand{\BIBdecl}{\relax}
\BIBdecl

\bibitem{Alexandre2016cvpr}
{A. Alahi, K. Goel, V. Ramanathan, A. Robicquet, L. Fei-Fei, S. Savarese},
  ``{Social LSTM: Human trajectory prediction in crowded spaces},'' \emph{{IEEE
  Conference on Computer Vision and Pattern Recognition (CVPR)}}, pp.
  {1063--6919}, {2016}.

\bibitem{Dehghan2018}
{A. Dehghan, M. Shah}, ``{Binary quadratic programing for online tracking of
  hundreds of people in extremely crowded scenes},'' \emph{{IEEE Transactions
  on Pattern Analysis And Machine Intelligence}}, vol.~{40}, no.~{3}, pp.
  {568--581}, {2018}.

\bibitem{Agrim2018}
{A. Gupta, J. Johnson, L. Fei-Fei, S. Savarese, A. Alahi}, ``{Social GAN:
  Socially acceptable trajectories with generative adversarial networks},''
  \emph{{IEEE/CVF Conference on Computer Vision and Pattern Recognition
  (CVPR)}}, pp. {2255--2264}, {2018}.

\bibitem{Kneidl2013}
{A. Kneidl, D. Hartmann, A. Borrmann}, ``{A hybrid multi-scale approach for
  simulation of pedestrian dynamics},'' \emph{{Transportation Research Part
  C-Emerging Technologies}}, vol.~{37}, pp. {223--237}, {2014}.

\bibitem{Lerner2007}
{A. Lerner, Y. Chrysanthou, D. Lischinski}, ``{Crowds by example},''
  \emph{{Computer Graphics Forum}}, vol.~{26}, pp. {1467--8659}, {2007}.

\bibitem{Alexandre2016}
{A. Robicquet, A. Sadeghian, A. Alahi, S. Savarese}, ``{Learning social
  etiquette: Human trajectory understanding in crowded scenes},''
  \emph{{European Conference on Computer Vision (ECCV) 2016, PT VIII}}, vol.
  {9912}, pp. {549--565}, {2016}.

\bibitem{Andrey2018}
{A. Rudenko, L. Palmieri, K. O. Arras}, ``{Joint long-term prediction of human
  motion using a planning-based social force approach},'' \emph{{IEEE
  International Conference on Robotics And Automation (ICRA)}}, pp.
  {4571--4577}, {2018}.

\bibitem{Vemula2017}
{A. Vemula, K. Muelling, J. Oh}, ``{Modeling cooperative navigation in dense
  human crowds},'' \emph{{IEEE International Conference on Robotics And
  Automation (ICRA)}}, pp. {1685--1692}, {2017}.

\bibitem{Bashar2018}
{B. I. Ahmad, J. K. Murphy, P. M. Langdon, S. J. Godsill}, ``{Bayesian intent
  prediction in object tracking using bridging distributions},'' \emph{{IEEE
  Transactions on Cybernetics}}, vol.~{48}, no.~{1}, pp. {215--227}, {2018}.

\bibitem{Pang2021}
{B. Pang, T. Zhao, X. Xie, Y. Wu}, ``{Trajectory prediction with latent belief
  energy-based model},'' \emph{{IEEE/CVF Conference on Computer Vision and
  Pattern Recognition (CVPR)}}, pp. {11\,809--11\,819}, {2021}.

\bibitem{Benjamin2016}
{B. Volz, K. Behrendt, H. Mielenz, I. Gilitschenski, R. Siegwart, J. Niet},
  ``{A data-driven approach for pedestrian intention estimation},'' \emph{{IEEE
  19th International Conference on Intelligent Transportation Systems (ITSC)}},
  pp. {2607--2612}, {2016}.

\bibitem{Barrios2015}
{C. Barrios, Y. Motai, D. Huston}, ``{Trajectory estimations using
  smartphones},'' \emph{{IEEE Transactions on Industrial Electronics}},
  vol.~{62}, no.~{12}, pp. {7901--7910}, {2015}.

\bibitem{Boucher2023}
{C. Boucher, R. Stower, V. S. Varadharajan, E. Zibetti, F. Levillain, D.
  St-Onge}, ``{Motion-based communication for robotic swarms in exploration
  missions},'' \emph{{Autonomous Robots}}, pp. {1--15}, {2023}, {DOI:
  10.1007/s10514-022-10079-0}.

\bibitem{David2020}
{D. Fridovich-Keil, A. Bajcsy, J. F. Fisac, S. L. Herbert, S. Wang, A. D.
  Dragan, C. J. Tomlin}, ``{Confidence-aware motion prediction for real-time
  collision avoidance},'' \emph{{The International Journal of Robotics
  Research}}, vol.~{39}, no. {2-3}, pp. {250--265}, {2020}.

\bibitem{Helbing1995}
{D. Helbing, P. Molnar}, ``{Social force model for pedestrian dynamics},''
  \emph{{Physical Review E}}, vol.~{51}, no.~{5}, pp. {4282--4286}, {1995}.

\bibitem{Lee2017}
{D. Lee, C. Liu, Y. Liao, J. K. Hedrick}, ``{Parallel interacting multiple
  model-based human motion prediction for motion planning of companion
  robots},'' \emph{{IEEE Transactions on Automation Science and Engineering}},
  vol.~{14}, no.~{1}, pp. {52--61}, {2017}.

\bibitem{Pool2017}
{Ewoud A. I. Pool, Julian F. P. Kooij, D. M. Gavrila}, ``{Using road topology
  to improve cyclist path prediction},'' \emph{{2017 IEEE Intelligent Vehicles
  Symposium (IV)}}, pp. {289--296}, {2017}.

\bibitem{Gers2000}
{F.A. Gers, J. Schmidhuber, F. Cummins}, ``{Learning to forget: Continual
  prediction with LSTM},'' \emph{{Neural Computation}}, vol.~{12}, no.~{10},
  pp. {2451--2471}, {2000}.

\bibitem{Golnaz2021}
{G. Habibi, J. P. How}, ``{Human trajectory prediction using similarity-based
  multi-model fusion},'' \emph{{IEEE Robotics and Automation Letters}},
  vol.~{6}, no.~{2}, pp. {715--722}, {2021}.

\bibitem{Aoude2013}
{G. S. Aoude, B. D. Luders, J. M. Joseph, N. Roy, J. P. How},
  ``{Probabilistically safe motion planning to avoid dynamic obstacles with
  uncertain motion patterns},'' \emph{{Autonomous Robots}}, vol.~{35}, no.~{1},
  pp. {51--76}, {2013}.

\bibitem{Xie2018}
{G. Xie, H. Gao, L. Qian, B. Huang, K. Li, J. Wang}, ``{Vehicle trajectory
  prediction by integrating physics- and maneuver-based approaches using
  interactive multiple models},'' \emph{{IEEE Transactions on Industrial
  Electronics}}, vol.~{65}, no.~{7}, pp. {5999--6008}, {2018}.

\bibitem{QiuH2017}
{H. Qiu, H. Duan}, ``{Pigeon interaction mode switch-based UAV distributed
  flocking control under obstacle environments},'' \emph{{ISA Transactions}},
  vol.~{71}, pp. {93--102}, {2017}.

\bibitem{Sun2022}
{H. Sun, Z. Zhao, Z. Yin, Z. He}, ``{Reciprocal twin networks for pedestrian
  motion learning and future path prediction},'' \emph{{IEEE Transactions on
  Circuits and Systems for Video Technology}}, vol.~{32}, no.~{3}, pp.
  {1483--1497}, {2022}.

\bibitem{WangH2021}
{H. Wang, B. Lu, J. Li, T. Liu, Y. Xing, C. Lv, D. Cao, J. Li, J. Zhang, E.
  Hashemi}, ``{Risk assessment and mitigation in local path planning for
  autonomous vehicles with LSTM based predictive model},'' \emph{{IEEE
  Transactions on Automation Science and Engineering}}, vol.~{19}, no.~{4}, pp.
  {2738--2749}, {2021}.

\bibitem{HuJY2022}
{J. Hu, A. E. Turgut, T. Krajnik, B. Lennox, F. Arvin}, ``{Occlusion-based
  coordination protocol design for autonomous robotic shepherding tasks},''
  \emph{{IEEE Transactions on Cognitive and Developmental Systems}}, vol.~{14},
  no.~{1}, pp. {126--135}, {2022}.

\bibitem{HU2022}
{J. Hu, B. Ding, M. Zhang, J. Zhao, Z. Xu, H. Pan}, ``{Enhancing output
  feedback robust MPC via lexicographic optimization},'' \emph{{IEEE
  Transactions on Industrial Informatics}}, vol.~{19}, no.~{3}, pp.
  {3068--3078}, {2023}.

\bibitem{Liang2020}
{J. Liang, L. Jiang, K. Murphy, T. Yu, A. Hauptmann}, ``{The garden of forking
  paths: Towards multi-future trajectory prediction},'' \emph{{2020 IEEE/CVF
  Conference on Computer Vision and Pattern Recognition (CVPR)}}, pp.
  {10\,505--10\,515}, {2020}.

\bibitem{Berg2011}
{J. v. d. Berg, S. J. Guy, M. Lin, D. Manocha}, ``{Reciprocal n-body collision
  avoidance},'' \emph{{Robotics Research}}, vol.~{70}, pp. {3--19}, {2011}.

\bibitem{Yamaguchi2011}
{K. Yamaguchi, A. C. Berg, L. E. Ortiz, T. L. Berg}, ``{Who are you with and
  where are you going?}'' \emph{{IEEE/CVF Conference on Computer Vision and
  Pattern Recognition (CVPR)}}, pp. {1345--1352}, {2011}.

\bibitem{Michael2015}
{M. Goldhammer, S. Koehler, K. Doll, B. Sick}, ``{Camera based pedestrian path
  prediction by means of polynomial least-squares approximation and multilayer
  perceptron neural networks},'' \emph{{SAI Intelligent Systems Conference}},
  pp. {390--399}, {2015}.

\bibitem{Lorente2018}
{M. T. Lorente, E. Owen, L. Montano}, ``{Model-based robocentric planning and
  navigation for dynamic environments},'' \emph{{The International Journal of
  Robotics Research}}, vol.~{37}, no.~{8}, pp. {867--889}, {2018}.

\bibitem{Helou2016}
{Rayan El Helou}, ``{Agent-based modelling of pedestrian microscopic
  interactions},'' Master's thesis, {Ohio State University}, {Columbus},
  {2016}.

\bibitem{Gaoshan2017}
{S. Gao, Q. Ye, J. Xing, A. Kuijper, Z. Han, J. Jiao, X. Ji}, ``{Beyond group:
  Multiple person tracking via minimal topology-energy-variation},''
  \emph{{IEEE Transactions on Image Processing}}, vol.~{26}, no.~{12}, pp.
  {5575--5589}, {2017}.

\bibitem{Kim2015}
{S. Kim, S. J. Guy, W. Liu, D. Wilkie, Rynson W. H. Lau, C. Lin, D. Manocha},
  ``{BRVO: Predicting pedestrian trajectories using velocity-space
  reasoning},'' \emph{{The International Journal of Robotics Research}},
  vol.~{34}, no.~{2}, pp. {201--217}, {2015}.

\bibitem{Mirjalili2017}
{S. Mirjalili, A. H. Gandomi, S. Z. Mirjalili, S. Saremi, H. Faris, S. M.
  Mirjalili}, ``{Salp swarm algorithm: A bio-inspired optimizer for engineering
  design problems},'' \emph{{Advances in Engineering Software}}, vol. {114},
  pp. {163--191}, {2017}.

\bibitem{Pellegrini2009}
{S. Pellegrini, A. Ess, K. Schindler, L. van Gool}, ``{You'll never walk alone:
  Modeling social behavior for multi-target tracking},'' \emph{{2009 IEEE 12th
  International Conference on Computer Vision}}, pp. {261--268}, {2009}.

\bibitem{Pellegrini2020}
{S. Pellegrini, A. Ess, L. Van Gool}, ``{Improving data association by joint
  modeling of pedestrian trajectories and groupings},'' \emph{{European
  Conference on Computer Vision (ECCV) 2010, PT I}}, vol. {6311}, pp.
  {452--465}, {2010}.

\bibitem{ChenY2022}
{Y. Chen, Y. Lou}, ``{A unified multiple-motion-mode framework for socially
  compliant navigation in dense crowds},'' \emph{{IEEE Transactions on
  Automation Science and Engineering}}, vol.~{19}, no.~{4}, pp. {3536--3548},
  {2022}.

\bibitem{Zhitian2021}
{Z. Zhang, J. Rhim, A. Lim, M. Chen}, ``{A multimodal and hybrid framework for
  human navigational intent inference},'' \emph{{IEEE/RSJ International
  Conference on Intelligent Robots and Systems (IROS)}}, pp. {993--1000},
  {2021}.

\end{thebibliography}
